\documentclass[10pt,twocolumn,letterpaper]{article}

\usepackage{cvpr}              

\usepackage{subfloat}
\usepackage{float}
\usepackage{graphicx}
\usepackage{caption} 
\usepackage{amsmath}
\usepackage{amssymb}
\usepackage{booktabs}
\usepackage{amsfonts,amssymb, amsthm, amsmath}
\usepackage{color}
\usepackage{rotating}
\usepackage{multirow}
\usepackage{multicol}
\usepackage{makecell}
\usepackage{diagbox}
\usepackage{soul}
\usepackage[accsupp]{axessibility}  
\usepackage[linesnumbered,ruled, vlined]{algorithm2e}
\usepackage{algpseudocode}

\usepackage[pagebackref,breaklinks,colorlinks]{hyperref}

\usepackage[capitalize]{cleveref}
\crefname{section}{Sec.}{Secs.}
\Crefname{section}{Section}{Sections}
\Crefname{table}{Table}{Tables}
\crefname{table}{Tab.}{Tabs.}

\newcommand{\Tref}[1]{Table~\ref{#1}}

\newcommand{\Fref}[1]{Figure~\ref{#1}}
\newcommand{\Sref}[1]{Section~\ref{#1}}

\def\cvprPaperID{3928} 
\def\confName{CVPR}
\def\confYear{2022}

\newcommand{\qiankun}[1]{\textcolor{black}{#1}}

\newcommand{\qchu}[1]{\textcolor{black}{#1}}

\begin{document}


\title{Reduce Information Loss in Transformers for Pluralistic Image Inpainting}

\author{Qiankun Liu\textsuperscript{1}\thanks{Work done during an internship at Microsoft} \quad Zhentao Tan\textsuperscript{1} \quad Dongdong Chen\textsuperscript{2} \quad Qi Chu\textsuperscript{1}\thanks{Corresponding author} \quad Xiyang Dai\textsuperscript{2} \\ 
Yinpeng Chen\textsuperscript{2} \quad Mengchen Liu\textsuperscript{2} \quad Lu Yuan\textsuperscript{2} \quad Nenghai Yu\textsuperscript{1} \\
\textsuperscript{1}CAS Key Laboratory of Electromagnetic Space Information, \\University of Science and Technology of China \\
\textsuperscript{2}Microsoft Cloud + AI \\
{\tt\small \{liuqk3, tzt\}@mail.ustc.edu.cn, \{qchu, ynh\}@ustc.edu.cn} \\
{\tt\small  cddlyf@gmail.com, \{xiyang.dai,  yiche, mengcliu, luyuan\}@microsoft.com}
}

\maketitle

\begin{abstract}
Transformers have achieved great success in pluralistic image inpainting recently.
However, we find existing transformer based solutions regard each pixel as a token, thus suffer from information loss issue from two aspects: 
1) They downsample the input image into much
lower resolutions for efficiency consideration, incurring information loss and extra misalignment 
for the boundaries of masked regions.
2) They quantize $256^3$ RGB pixels to a small number (such as 512) of quantized pixels. The indices of quantized pixels are used as tokens for the inputs and prediction targets of transformer.
Although an extra CNN network is used to upsample and refine the low-resolution results, it is difficult to retrieve the lost information back.
\qchu{\qiankun{To keep input information as much as possible, we propose a new transformer based framework ``\textit{\textbf{PUT}}''.} 
Specifically, to avoid input downsampling while maintaining the computation efficiency, we design a patch-based auto-encoder \textbf{P}-VQVAE, where the encoder converts the masked image into non-overlapped patch tokens and the decoder recovers the masked regions from inpainted tokens while keeping the unmasked regions unchanged. To eliminate the information loss caused by quantization, an Un-Quantized Transformer (\textbf{U}Q-\textbf{T}ransformer) is applied, which directly takes the features from P-VQVAE encoder as input without quantization and regards the quantized tokens only as prediction targets.}
Extensive experiments show that PUT greatly outperforms state-of-the-art methods on image fidelity, especially for large masked regions and complex large-scale datasets. Code is available at \url{https://github.com/liuqk3/PUT}
\end{abstract}
\vspace{-20pt}

\section{Introduction}
\label{sec:intro}
    Image inpainting, which focuses on filling meaningful and plausible contents in missing regions for the damaged images, has always been a hot topic in computer vision areas and widely used in various applications \cite{sun2018moire, qiu2020semanticadv, zhan2020self, barnes2009patchmatch,wan2020bringing,wan2022old,tan2020michigan}. Traditional methods \cite{bertalmio2003simultaneous, barnes2009patchmatch, criminisi2004region} based on texture matching can handle simple cases very well but struggle for complex natural images. In the last several years, benefiting from development of CNNs, tremendous success \cite{pathak2016context,liu2018image,liu2019coherent,yu2018generative} has been achieved by learning on large-scale datasets. However, due to the inherent properties of CNNs, i.e., local inductive bias and spatial-invariant kernels, such methods still do not perform  well in understanding global structure and inpainting large masked/missing regions. 
   
	\begin{figure}[t]
		\centering
		\includegraphics[width=1.0\columnwidth]{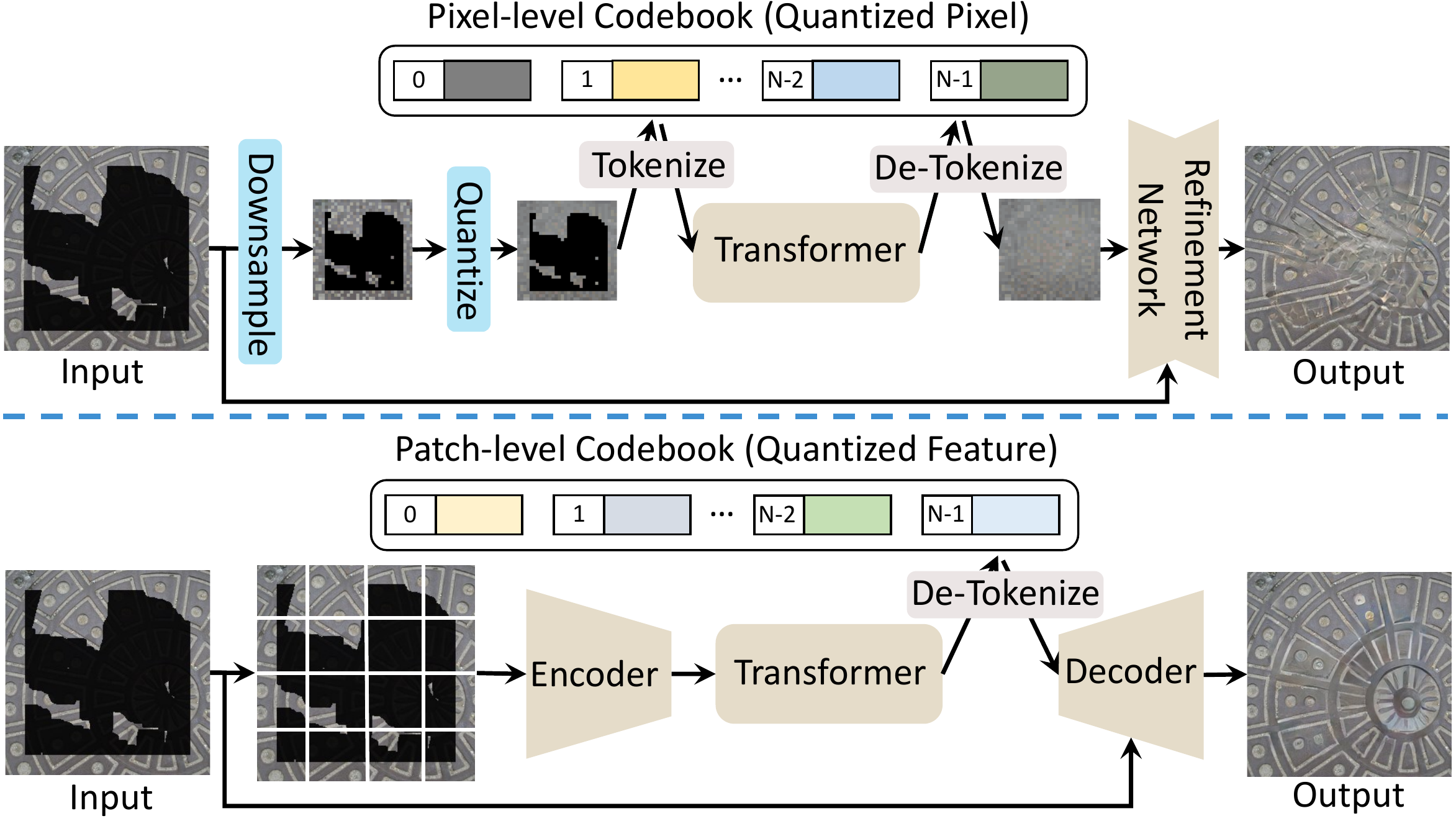} 
		\caption{Top: Existing transformer based methods \cite{wan2021high}.
		The output is produced by ICT \cite{wan2021high}. 
		Bottom: Our transformer based method. 
		\emph{``Tokenize''} here means getting the indices of quantized pixels or features, and \emph{``De-Tokenize''} is the inverse operation.
		}
		\vspace{-8pt}
		\label{figure: different_type_of_methods}
	\end{figure}

    Recently, transformers have demonstrated their power in various vision tasks \cite{carion2020end,dong2021cswin,chen2022mobile,chen2021pix2seq, chen2020generative, esser2021taming, ramesh2021zero, chen2021transformer, wang2021transformer}, thanks to their capability of modeling long-term relationship. Some recent works \cite{wan2021high} also attempt to apply transformers for pluralistic image inpainting, and have achieved remarkable success in better diversity and large region inpainting quality. As shown in the top row of \Fref{figure: different_type_of_methods}, they follow the similar design: 1) Downsample the input image into lower resolutions and quantize the pixels; 2) Use the transformer to recover the masked pixels by regarding each quantized pixel as the token; 3) Upsample and refine the low-resolution result by feeding it together with the original input image into an extra CNN network.

    In this paper, we argue that using the above pixel-based token makes existing transformer based solutions suffer from the information loss issue from two aspects: 1) \emph{``Low resolution"}. To avoid high computation complexity of transformer, the input image is downsampled into much lower resolution to reduce the input token number, which not only incurs information loss but also introduces misalignment 
    for the boundaries of masked regions when upsampled back to the original resolution. 
    2) \emph{``Quantization"}. To constrain the prediction within a small space, the huge amount ($256^3$, in detail) of RGB pixels are quantized into much less (such as 512) qunatized pixels through clustering. The indices of quantized pixels are used as discrete tokens both for the input and prediction target of transformer. Such practice would further result in the information loss.

 \begin{figure}[t]
	\centering
	\includegraphics[width=1.0\columnwidth]{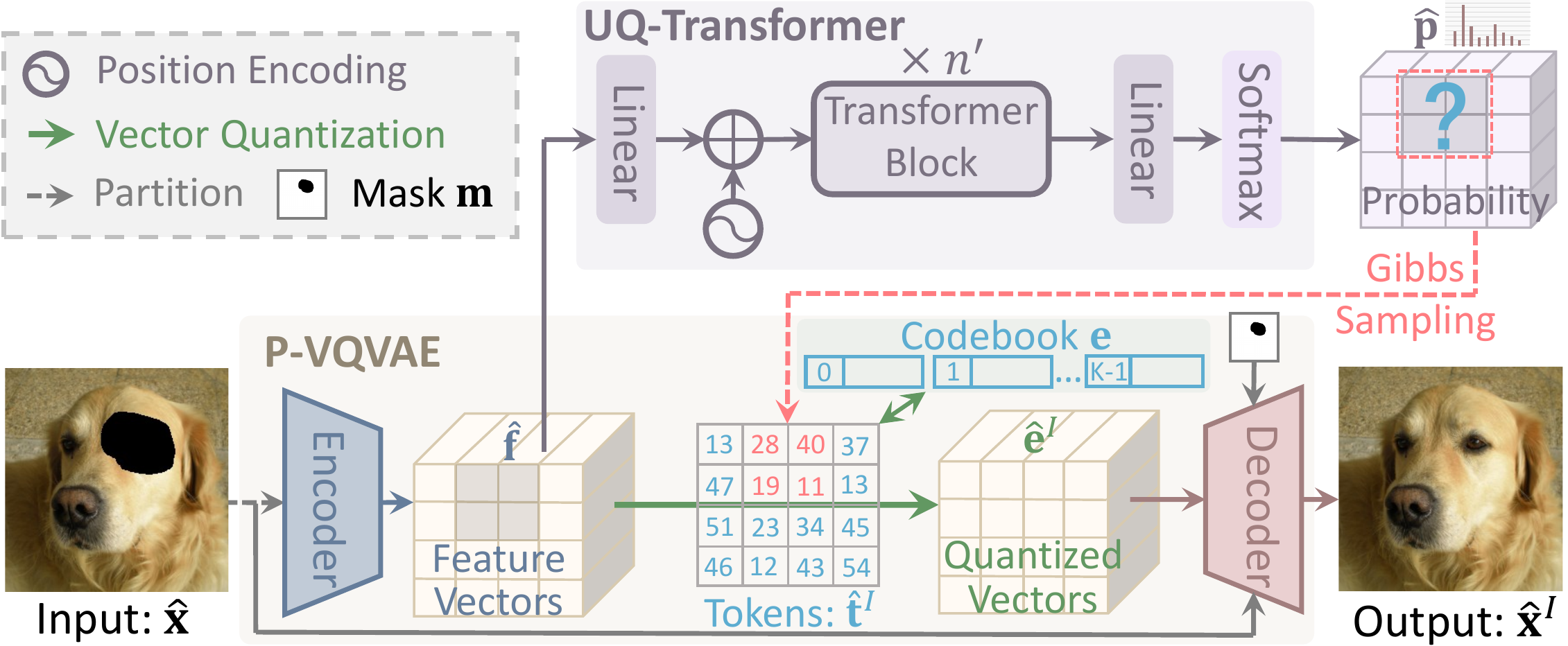} 
	\caption{Pipeline of PUT for pluralistic image inpainting. 
	Note that the pluralistic inpainting results are not shown in the figure.
	}
	\vspace{-8pt}
	\label{figure: transformer_framework}
\end{figure}

    \qchu{To mitigate these issues, we propose a new transformer based framework \textbf{PUT}, which can reduce the information loss as much as possible. As shown in the bottom row of \Fref{figure: different_type_of_methods}, the original high-resolution input image is directly fed into a patch based encoder without any downsampling and the transformer directly takes the features from the encoder as input without any quantization. Specifically, \textbf{PUT} contains two key designs: Patch-based Vector Quantized Variational Auto-Encoder (``\textit{P-VQVAE}", \Sref{sec: p_vqvae}) and Un-Quantized Transformer(``\textit{UQ-Transformer}", \Sref{sec: ut_transformer}).} 
    \textit{P-VQVAE} is a specially designed patch auto-encoder: 1) Its encoder converts each image patch into the latent feature in a non-overlapped way, where the non-overlapped design is to avoid the disturbance between masked regions and unmasked regions; 2) As the prediction space of UQ-Transformer, a dual-codebook is built for patch feature tokenization, where masked patches and unmasked patches are separately represented by 
    different codebooks;
    3) The decoder in P-VQVAE not only recovers the masked image regions from the inpainted tokens but also maintains unmasked regions unchanged. For \textit{UQ-Transformer}, it utilizes the quantized tokens of unmasked patches as the prediction targets for masked patches, but takes the un-quantized feature vectors from the encoder as input. Compared to taking the quantized tokens as input, this design can avoid the information loss and help UQ-Transformer make more accurate predictions.
    
    To demonstrate the superiority, we conduct extensive experiments on FFHQ \cite{karras2019style}, Places2 \cite{zhou2017places} and ImageNet \cite{deng2009imagenet}. The results show that our method outperforms CNN based pluralistic inpainting methods by a large margin on different evaluation metrics. Benefiting from less information loss, our method also achieves much higher fidelity than existing transformer based solutions, especially for large region inpainting and complex large-scale datasets.

\begin{figure*}[t]
	\centering
	\includegraphics[width=2.1\columnwidth]{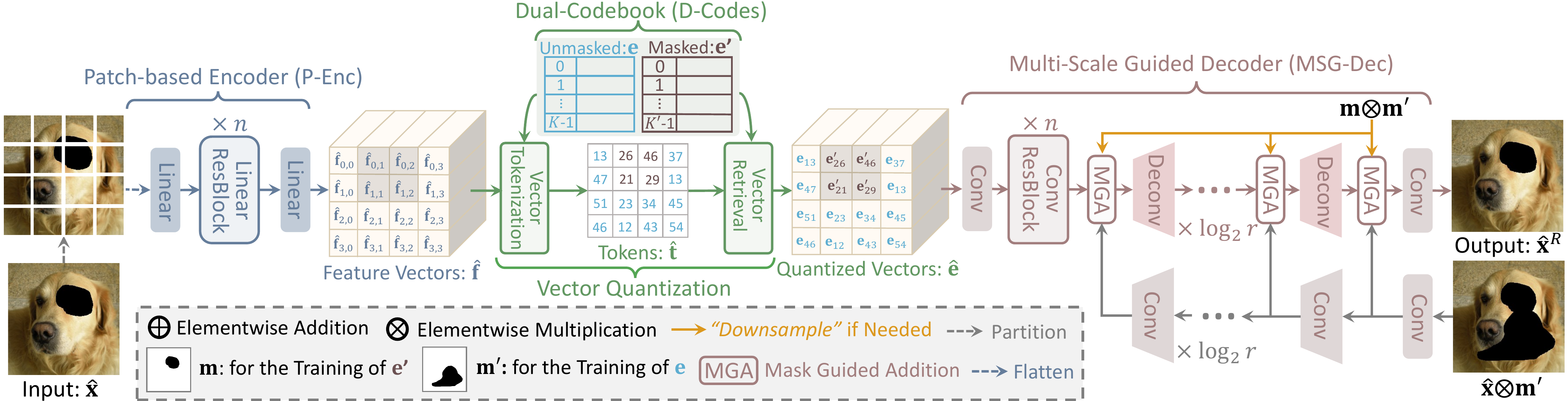} 
	\caption{Training procedure of P-VQVAE. 
	The detailed architecture of P-VQVAE could be found in the supplementary material.
	}
	\vspace{-8pt}
	\label{figure: p_vqvae_framework}
\end{figure*}

\section{Related Work}
\label{sec:related work}

\noindent\textbf{Auto-Encoders.}
Auto-encoders \cite{hinton1994autoencoders} is a kind of artificial neural network in semi-supervised and unsupervised learning. Among its subclasses, variational auto-encoders (VAE) \cite{kingma2013auto,doersch2016tutorial} is widely used for image synthesis tasks \cite{tan2021efficient,tan2021diverse} as a generative model. 
It can be trained in self-supervised strategy and generate diverse images through decoder with latent space sampling or autoregressive models \cite{van2016pixel,oord2016conditional}. Later, vector quantized variational auto-encoder (VQ-VAE) \cite{van2017neural} is proposed for discrete representation learning to circumvent issues of ``posterior collaps'', and further developed by VQ-VAE-2 \cite{razavi2019generating}. Recently, based on the similar quantization mechanism with VQ-VAE, VAGAN \cite{esser2021taming} and dVAE \cite{ramesh2021zero} are proposed for conditonal image generation through transformers \cite{vaswani2017attention}, while PeCo \cite{dong2021peco} trains a perceptual vision tokenizer for vision transformer BERT-pretraining \cite{wang2022bevt}. Different from previous methods, the proposed ``\textit{P-VQVAE}", which contains a non-overlapped patch encoder, a dual-codebook and 
a multi-scale guided decoder, is dedicated for image inpainting.

\vspace{0.2em}
\noindent\textbf{Visual Transformers.}
Thanks to the capability of long range relationship modeling, transformers have been widely used in different vision tasks, such as object detection \cite{carion2020end, chen2021pix2seq},  image synthesis \cite{chen2020generative, esser2021taming, ramesh2021zero}, object tracking \cite{chen2021transformer, wang2021transformer} and image inpainting \cite{wan2021high}. 
Specifically, the autoregressive inference mechanism is naturally suitable for image synthesis related tasks, which can bring diverse results while guarantee the quality of the synthesized images \cite{chen2020generative,esser2021taming,ramesh2021zero,wan2021high}. In this paper, we take full advantage of transformer and propose to replace discrete tokens with continuous feature vectors to avoid the information loss.

\vspace{0.2em}
\noindent\textbf{Image Inpainting.}
According to the diversity of inpainted images, there are two different types of definition for image inpainting task: deterministic image inpainting and pluralistic image inpainting. Most of the traditional methods, whether diffusion-based methods \cite{bertalmio2000image,efros2001image} or path-based methods \cite{barnes2009patchmatch,hays2007scene,darabi2012image}, can only generate single result for each input and may be failed while meeting large-area of missing pixels. Later, some CNN based methods \cite{pathak2016context,liu2018image,liu2019coherent,yu2018generative,nazeri2019edgeconnect,iizuka2017globally} are proposed to ensure consistency of the semantic content of the inpainted images, but still ignore the diversity of results. To generate several diverse results for each masked image,  
some CNN based \cite{zheng2019pluralistic,zhao2020uctgan} and transformer based  \cite{wan2021high} methods have emerged recently. Among them, transformer based methods \cite{wan2021high} show their merits in both quality and diversity than CNN based methods.
However, their unreasonable design, like downsampling of input image and quantization of transformer inputs, results in the serious information loss issue.
Thus, we propose a novel framework, PUT, which maximizes the input information to achieve better synthesis results.

\section{Method}
\label{sec: method}
The proposed method mainly consists of a Patch-based Vector Quantized Variational Auto-Encoder (P-VQVAE) and an Un-Quantized transformer (UQ-Transformer). The overview of our method is shown in \Fref{figure: transformer_framework}. Let ${\mathbf x} \in \mathbb{R}^{H \times W \times 3}$ be an image and ${\mathbf m} \in \{0,1\}^{H \times W \times 1}$ be the mask denoting whether a region needs to be inpainted (with value 0) or not (with value 1). $H$ and $W$ are the spatial resolution. The image $\mathbf{\hat{x}}=\mathbf{x} \otimes \mathbf{m}$ is the masked image that contains missing pixels, where $\otimes$ is the elementwise multiplication. 
The masked image $\mathbf{\hat{x}}$ is first fed into the encoder of P-VQVAE to get the patch based feature vectors. Then UQ-Transformer takes the feature vectors as input and predicts the tokens (i.e., indices) of latent vectors in a codebook for masked regions. 
Finally, the retrieved latent vectors are used as the quantized vectors for patches
and fed into the decoder of P-VQVAE to reconstruct the inpainted image.

\subsection{P-VQVAE}
\label{sec: p_vqvae}
\qchu{To avoid the information loss from input downsampling while maintaining the computation efficiency for the transformer, we utilize the merits of auto-encoder to replace the downsampled pixels with the features from encoder. Compared to the downsampled pixels, the features from encoder can have the same low-resolution  for efficiency while contains more information for reconstruction. Considering the task of image inpainting, we specially design P-VQVAE, 
which contains a patch-based encoder, a dual-codebook and a multi-scale guided decoder.}

\paragraph{Patch-based Encoder.}
\label{sec: p_enc}
\qchu{Conventional CNN based encoders process the input image with several convolution kernels in a sliding window manner, which are unsuitable for image inpainting since they would introduce disturbance between masked and unmasked regions. Thus, the encoder of P-VQVAE (denoted as P-Enc) is designed to process input image by several linear layers in a non-overlapped patch manner.}
Specifically, the masked image $\mathbf{\hat{x}}$ is firstly partitioned into $\frac{H}{r} \times \frac{W}{r}$ non-overlapped patches, where $r$ is the spatial size of patches 
and is set to 8 by default. For a patch, we call it a \emph{masked patch} if it contains any missing pixels, otherwise \emph{unmasked patch}.
Each patch is flattened and then mapped into a feature vector.
Formally, all feature vectors are denoted as $\mathbf{\hat{f}} = \mathcal{E}(\mathbf{\hat{x}})\in \mathbb{R}^{\frac{H}{r}\times\frac{W}{r}\times C}$, where $C$ (set to 256 by default) is the dimensionality of feature vectors and $\mathcal{E}(\cdot)$ is the encoder function. 

\paragraph{Dual-Codebook for Vector Quantization.}
\label{sec: d_codes}
Following the works in \cite{van2017neural, razavi2019generating, esser2021taming}, the feature vectors from encoder are quantized into discrete tokens with the latent vectors in the learnable codebook. By contrast, we design a dual-codebook (denote as D-Codes) for vector quantization, which is more suitable for image inpainting. In D-Codes, the latent vectors are divided into two parts, denoted as $\mathbf{e} \in \mathbb{R}^{K\times C}$ and $\mathbf{e'}\in \mathbb{R}^{K'\times C}$, which are responsible for feature vectors that mapped from unmasked and masked patches respectively. $K$ and $K'$ are the number of latent vectors.
Let $\mathbf{m}^{\downarrow} \in \{0,1\}^{\frac{H}{r}\times \frac{W}{r} \times 1}$ be the indicator mask that indicates whether a patch is a masked (with value 0) or unmasked (with value 1) patch.
The feature vector $\mathbf{\hat{f}}_{i,j}$ is quantized as:%
\begin{equation}
	\begin{cases}
		\mathbf{e}_k \ {\rm where} \ k = {\rm argmin}_{l}\parallel \mathbf{\hat{f}}_{i,j} \ominus \mathbf{e}_l \parallel_2, \ {\rm if} \ \mathbf{m}^{\downarrow}_{i,j} = 1, \\
		\mathbf{e'}_{k'} \ {\rm where} \ k' = {\rm argmin}_{l} \parallel \mathbf{\hat{f}}_{i,j} \ominus \mathbf{e'}_l \parallel_2, \ {\rm else},
	\end{cases}
	\label{eq: vector_quantization}
\end{equation}
where $\ominus$ denotes the operation of elementwise subtraction.  
Let $\mathbf{\hat{e}} \in \mathbb{R}^{\frac{H}{r} \times \frac{W}{r} \times C}$ and $\mathbf{\hat{t}} = \mathcal{I}(\mathbf{\hat{f}}, \mathbf{e}, \mathbf{e'}, \mathbf{m}^{\downarrow}) \in \mathbb{N}^{\frac{H}{r} \times \frac{W}{r}}$ be the quantized vectors and tokens for $\mathbf{\hat{f}}$, where $\mathcal{I}(\cdot, \cdot, \cdot, \cdot)$ denotes the function that gets tokens for its first argument and it can be simply implemented by getting the indices of all quantized vectors in $\mathbf{\hat{e}}$.
The dual-codebook helps P-Enc learn more discriminative feature vectors for masked and unmasked patches since they are quantized and represented with different codebooks, which further disenchants transformer about the masked and unmasked patches to predict more reasonable results for masked patches.

\paragraph{Multi-Scale Guided Decoder.}
\label{sec: msg_dec}
For image inpainting task, an indisputable fact is that the unmasked regions should be kept unchanged. To this end, we design a multi-scale guided decoder, denoted as MSG-Dec, to construct the inpainted image $\mathbf{\hat{x}}^{I}$ by referencing the input masked image $\mathbf{\hat{x}}$. 
Let $\mathbf{\hat{t}}^{I}$ be the inapinted tokens by transformer (Ref. \Fref{figure: transformer_framework} and \Sref{sec: sampling_strategy}) and $\mathbf{\hat{e}}^{I}$ be the retrieved quantized vectors from codebook based on $\mathbf{\hat{t}}^I$. The construction procedure is formulated as:
\begin{equation}
	\mathbf{\hat{x}}^{I} = \mathcal{D}(\mathbf{\hat{e}}^{I}, \mathbf{m}, \mathbf{\hat{x}}),
	\label{eq: construction_of_inpainted_image}
\end{equation}
where $\mathcal{D}(\cdot, \cdot, \cdot)$ is the decoder function. The decoder consists of two branches: a main branch which starts with the quantized vectors $\mathbf{\hat{e}}^{I}$ and uses several deconvolution layers to generate inpainted images and a reference branch which extracts multi-scale feature maps (with spatial sizes $\frac{H}{2^{l}} \times \frac{W}{2^{l}}$, $0 \le l \le {\rm log_2}r$) from the masked image $\mathbf{\hat{x}}$. The features from the reference branch are fused to the features with the same scale from the main branch through a Mask Guided Addition (MGA) module as:
\begin{equation}
	\mathbf{\hat{e}}^{I,l-1} = {\rm Deconv}( (1-\mathbf{m}^{\downarrow,l}) \otimes \mathbf{\hat{e}}^{I,l} + \mathbf{m}^{\downarrow,l} \otimes \mathbf{\hat{f}}^{R,l}),
	\label{eq: mask_guided_addition}
\end{equation}
where $\mathbf{\hat{e}}^{I,l}$ and $\mathbf{\hat{f}}^{R,l}$ are the features with spatial size $\frac{H}{2^{l}} \times \frac{W}{2^{l}}$ from the main branch and the reference branch respectively. $\mathbf{m}^{\downarrow,l}$ is the indicator mask obtained from $\mathbf{m}$ for corresponding spatial size.

\paragraph{Training of P-VQVAE.}
To avoid the decoder learning to reconstruct input image $\mathbf{\hat{x}}$ only from the reference image, we get the reference image by randomly erasing some pixels in $\mathbf{\hat{x}}$ with another mask $\mathbf{m'}$ (see in \Fref{figure: p_vqvae_framework}). Let $\mathbf{\hat{x}}^R = \mathcal{D}(\mathbf{\hat{e}}, \mathbf{m} \otimes \mathbf{m'}, \mathbf{\hat{x}} \otimes \mathbf{m'})$ be the reconstructed image. In our design, the unmasked pixels in the reference image will be utilized to recover the corresponding pixels in $\mathbf{\hat{x}}^R$, while the latent vectors in codebook $\mathbf{e}'$ and $\mathbf{e}$ will be used to recover the pixels in $\mathbf{\hat{x}}^R$ masked by $\mathbf{m}$ and the remaining pixels respectively. The loss for training P-VQVAE is:
\begin{equation}
\begin{aligned}
     L_{vae}= \mathcal{L}_{rec}(\mathbf{\hat{x}}, \mathbf{\hat{x}}^R) 
	 + \parallel {\rm sg}[\mathbf{\hat{f}}] \ominus \mathbf{\hat{e}} \parallel^2_2 
	 + \beta \parallel {\rm sg}[\mathbf{\hat{e}}] \ominus \mathbf{\hat{f}} \parallel^2_2,
\end{aligned}
\label{eq: p_vqvae_loss}
\end{equation}
where 
${\rm sg}[\cdot]$ refers to a stop-gradient operation that blocks gradients from flowing into its argument.
$\beta$ is the weight for balance and is set to 0.25. $\mathcal{L}_{rec}(\cdot, \cdot)$ is the function to measure the difference between inputted and reconstructed images, including the L1 loss between the pixel values in two images and the gradients of two images, the adversarial loss\cite{goodfellow2014generative} obtained by a discriminator network, as well as the perceptual loss \cite{johnson2016perceptual} and the style loss\cite{gatys2016image}  between the two images.
Following \cite{van2017neural,razavi2019generating}, the second term in Eq. (\ref{eq: p_vqvae_loss}) is replaced by Exponential Moving Average (EMA) to optimize the vectors in D-Codes. More details about the training of P-VQVAE could be found in the supplementary material.

\begin{figure*}[t]
	\centering
	\includegraphics[width=2.07\columnwidth]{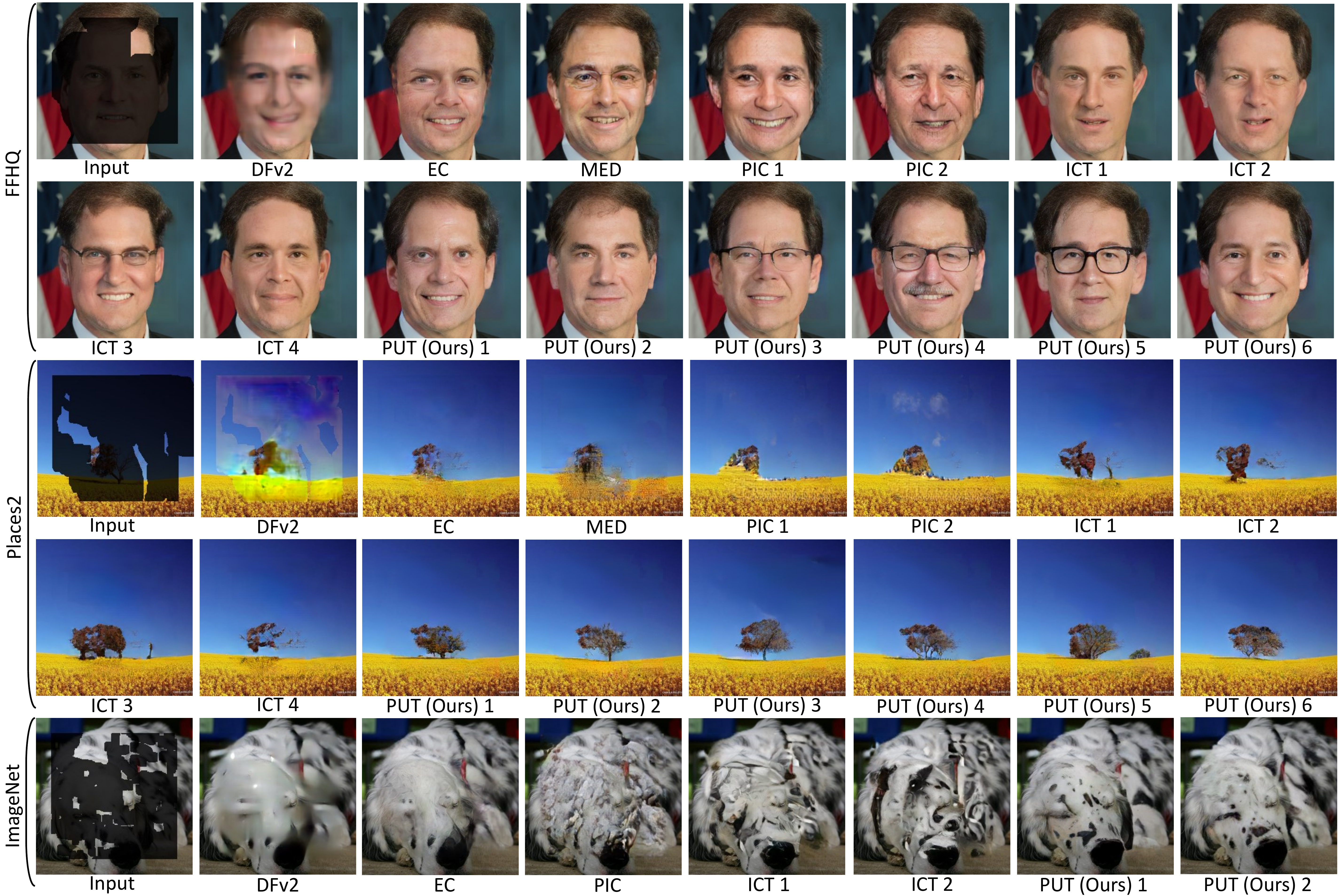} 
    \vspace{-5pt}
	\caption{Samples of inpainted results produced by different methods. For PUT, we set $\mathcal{K}=50$. More qualitative comparisons are presented in the supplementary material.}
	\vspace{-8pt}
	\label{figure: inpainting_results}
\end{figure*}

\subsection{UQ-Transformer}
\label{sec: ut_transformer}
In existing transformers for image inpainting \cite{wan2021high} and synthesis \cite{esser2021taming, ramesh2021zero}, the quantized discrete tokens are used as both the inputs and prediction targets. Given such discrete tokens, transformers
suffer from the severe information loss issue, which is harmful to their prediction.
In contrast, to take full advantage of feature vectors $\mathbf{\hat{f}}$ from the encoder of P-VQVAE, our UQ-Transformer directly takes them as the inputs and predicts the discrete tokens for masked patches.

Specifically, $\mathbf{\hat{f}}$ is firstly mapped by a linear layer and then added with extra learnable position embeddings for the encoding of spatial information. Finally, following \cite{radford2019language}, the feature vectors are flattened along spatial dimension to get the final input for the subsequent several transformer blocks.
The output of the last transformer block is further projected to the distribution over $K$ latent vectors in codebook $\mathbf{e}$ with a linear layer and a softmax function. We formulate the above procedure as $\mathbf{\hat{p}} = \mathcal{T}(\mathbf{\hat{f}}) \in [0, 1]^{\frac{H}{r}\times \frac{W}{r} \times K}$, where $\mathcal{T}(\cdot)$ refers the UQ-Transformer function.

\paragraph{Training of UQ-Transformer.}
Given a masked image $\mathbf{\hat{x}}$, the distribution of its corresponding inpainted tokens over $K$ latent vectors can be obtained with the pre-trained P-VQVAE and UQ-Transformer $\mathbf{\hat{p}} = \mathcal{T}(\mathcal{E}(\mathbf{\hat{x}}))$.
The ground-truth tokens for $\mathbf{x}$ is $\mathbf{t} = \mathcal{I}(\mathcal{E}(\mathbf{x}), \mathbf{e}, \mathbf{e'}, \mathcal{O}(\mathbf{m}^{\downarrow}))$  (Ref. \Sref{sec: d_codes}), where $\mathcal{O}(\cdot)$ sets all values in the given argument to 1. 
UQ-Transformer is trained with cross-entropy loss by fixing P-VQVAE:
\begin{equation}
\begin{aligned}
    L_{trans} = \frac{-1}{\sum_{i,j} 1 - \mathbf{m}_{i,j}^{\downarrow}} \sum_{i,j} (1-\mathbf{m}^{\downarrow}_{i,j}) {\rm log} \mathbf{\hat{p}}_{i,j,\mathbf{t}_{i,j}}.
\end{aligned}
\label{eq: ut_transformer_loss}
\end{equation}
In order to make the training stage consistent with inference stage, where only the quantized vectors can be obtained for masked regions, we randomly quantize the feature vectors in $\mathcal{E}(\mathbf{\hat{x}})$ to the latent vectors in codebook with probability 0.3 before feeding it to UQ-Transformer. 

\begin{figure}[t]
	\centering
	\includegraphics[width=1.0\columnwidth]{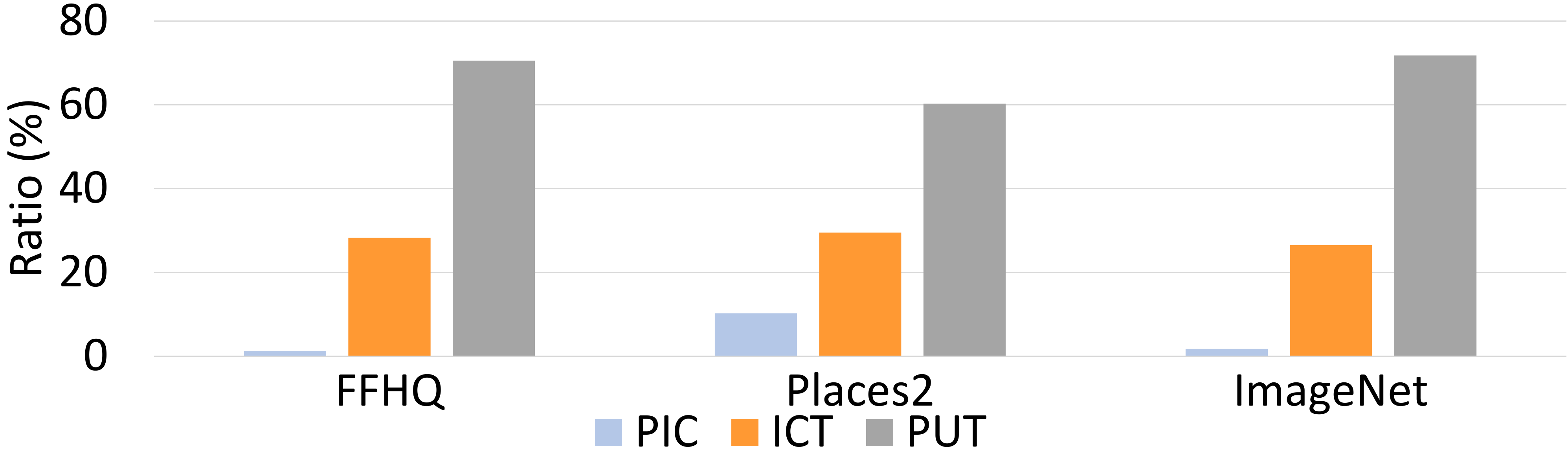} 
    \vspace{-10pt}
    \caption{The ratio of each method among the rank 1 images evaluated by human. Statistics are collected from 23 participants.}
    \vspace{-10pt}
	\label{figure: usr_study}
\end{figure}

\subsection{Sampling Strategy for Image Inpaining}
\label{sec: sampling_strategy}
For the production of diverse results, the tokens for masked patches ($\mathbf{m}_{i,j}^{\downarrow}=0$) are iteratively sampled with Gibbs sampling. Specifically, in each iteration, we first select the patch with the maximum predicted probability among the remaining masked patches. Then the token for the selected patch is sampled from the top-$\mathcal{K}$ predicted elements. Finally, the corresponding latent vector for the sampled token is retrieved to replace the feature vector of the selected patch before feeding UQ-Transformer for the next iteration. After sampling the tokens for all masked patches, we can get all quantized vectors $\mathbf{\hat{e}}^I$ with the inpainted tokens $\mathbf{\hat{t}}^I$, and the inpainted image can be constructed using Eq. (\ref{eq: construction_of_inpainted_image}). 
For the production of deterministic results, the tokens for masked patches are sampled with the largest probabilities at once.

\section{Experiments}
\label{sec: experiments}
The evaluation is conducted at $256 \times 256$ (i.e., $H=256$ and $W=256$) resolution on three different datasets, including FFHQ \cite{karras2019style}, Places2 \cite{zhou2017places} and ImageNet \cite{deng2009imagenet}. We use the original training and testing splits for Places2 and ImageNet. For FFHQ, we maintain the last 1K images for evaluation and others for training. Following ICT \cite{wan2021high}, only 1K images are randomly chosen from the test split of ImageNet for evaluation and the irregular masks provided by PConv \cite{liu2018image} are used both for training and testing.

\begin{figure}[t]
	\centering
	\subfloat[FID on FFHQ \cite{karras2019style}]{
		\begin{minipage}[t]{0.5\linewidth}
			\flushleft
			\includegraphics[width=0.95\linewidth]{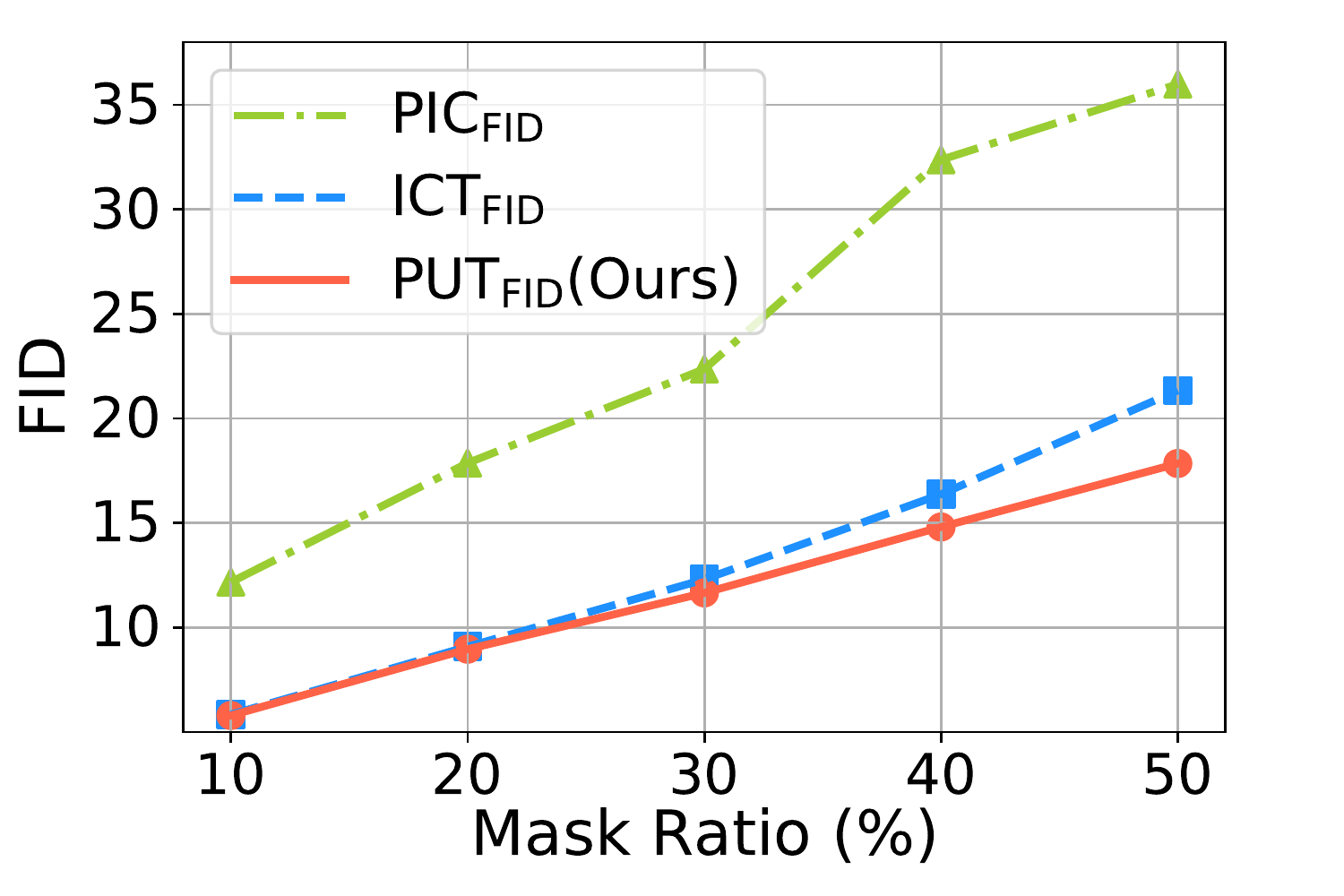}
		\end{minipage}%
	}%
	\subfloat[LPIPS on FFHQ \cite{karras2019style}]{
		\begin{minipage}[t]{0.5\linewidth}
			\centering
			\includegraphics[width=0.99\linewidth]{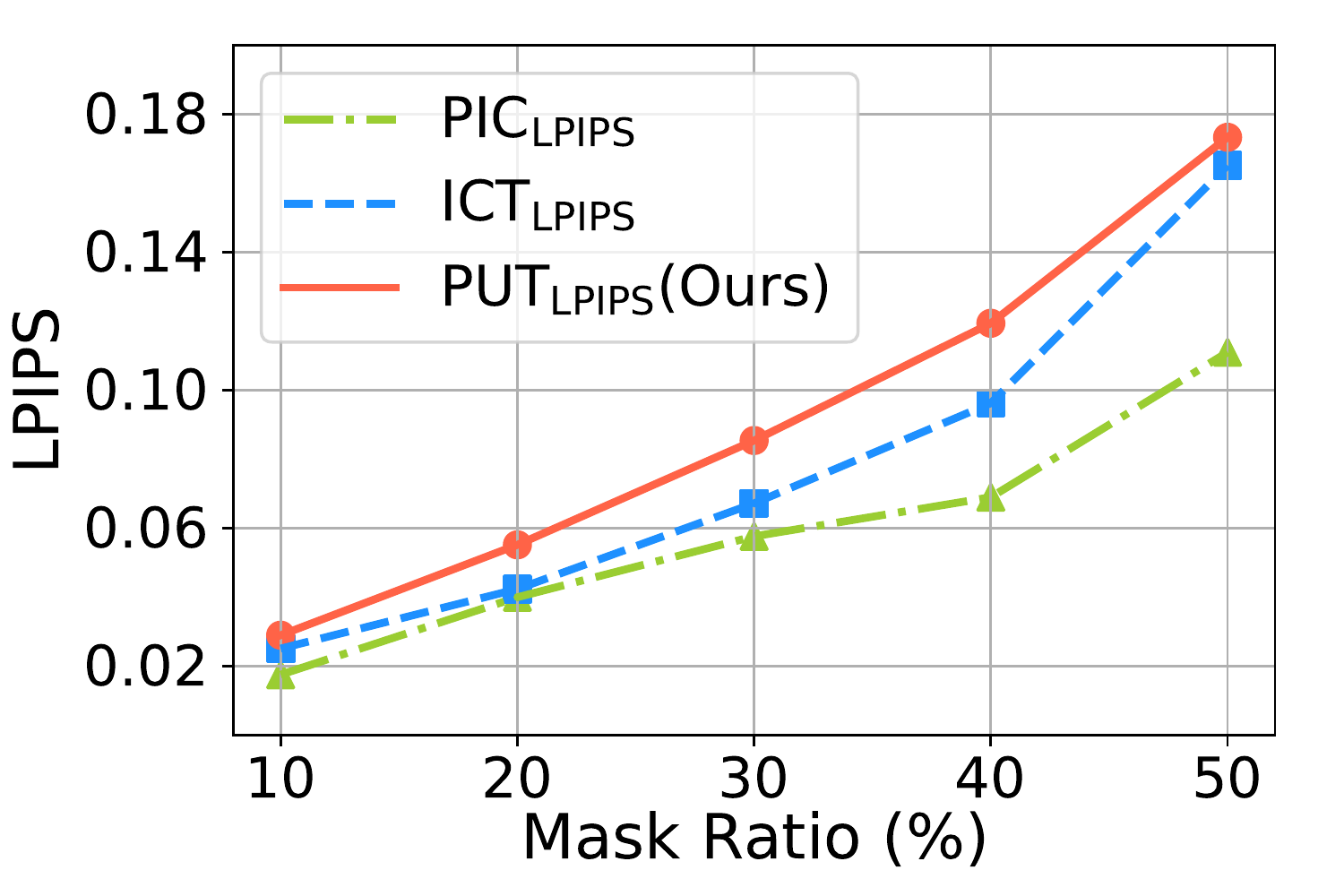}
		\end{minipage}%
	}%
	
	\subfloat[FID on Places2 \cite{zhou2017places}]{
		\begin{minipage}[t]{0.5\linewidth}
            \flushleft
			\includegraphics[width=0.95\linewidth]{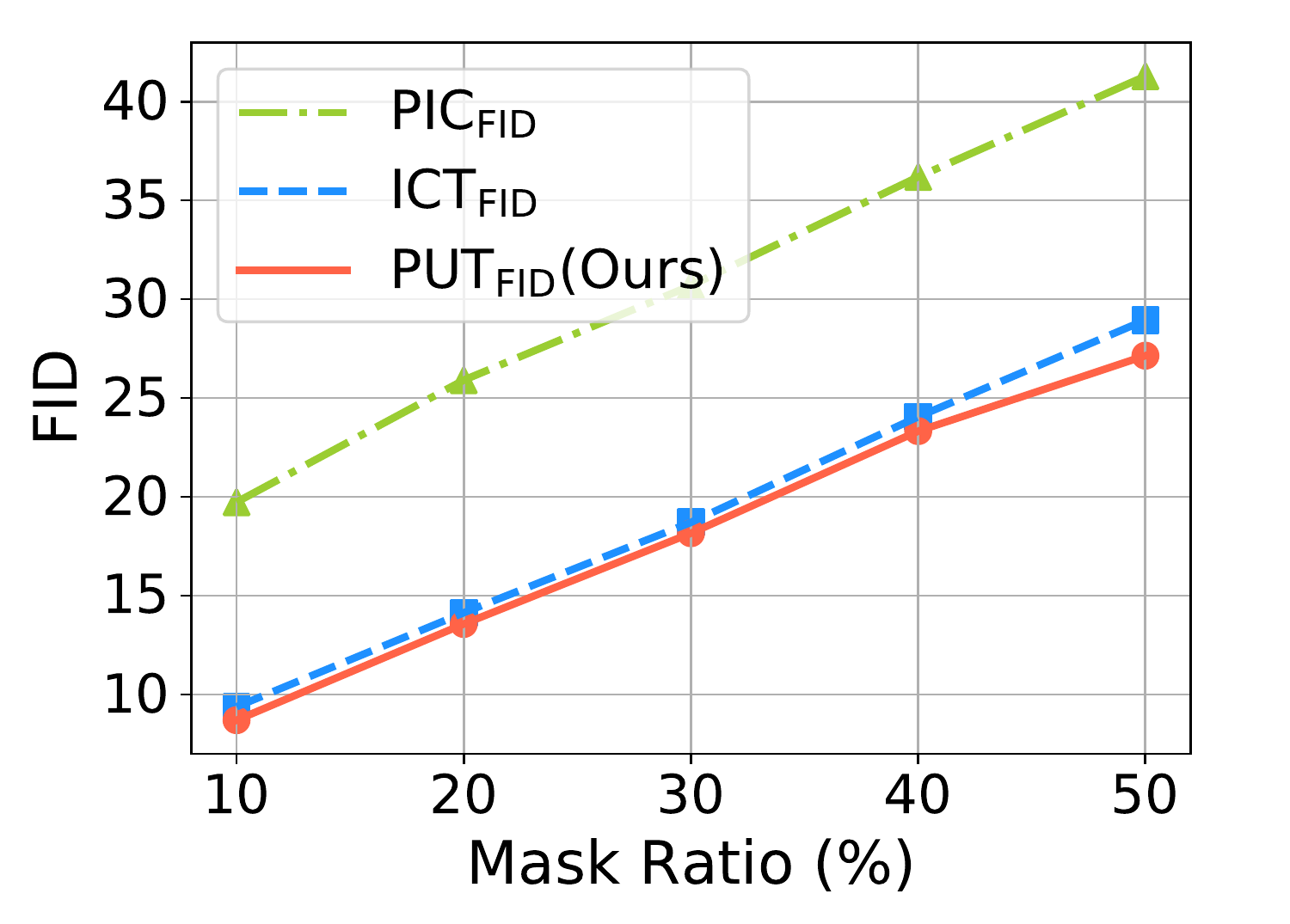}
		\end{minipage}
	}%
	\subfloat[LPIPS on Places2 \cite{zhou2017places}]{
		\begin{minipage}[t]{0.5\linewidth}
			\centering
			\includegraphics[width=0.99\linewidth]{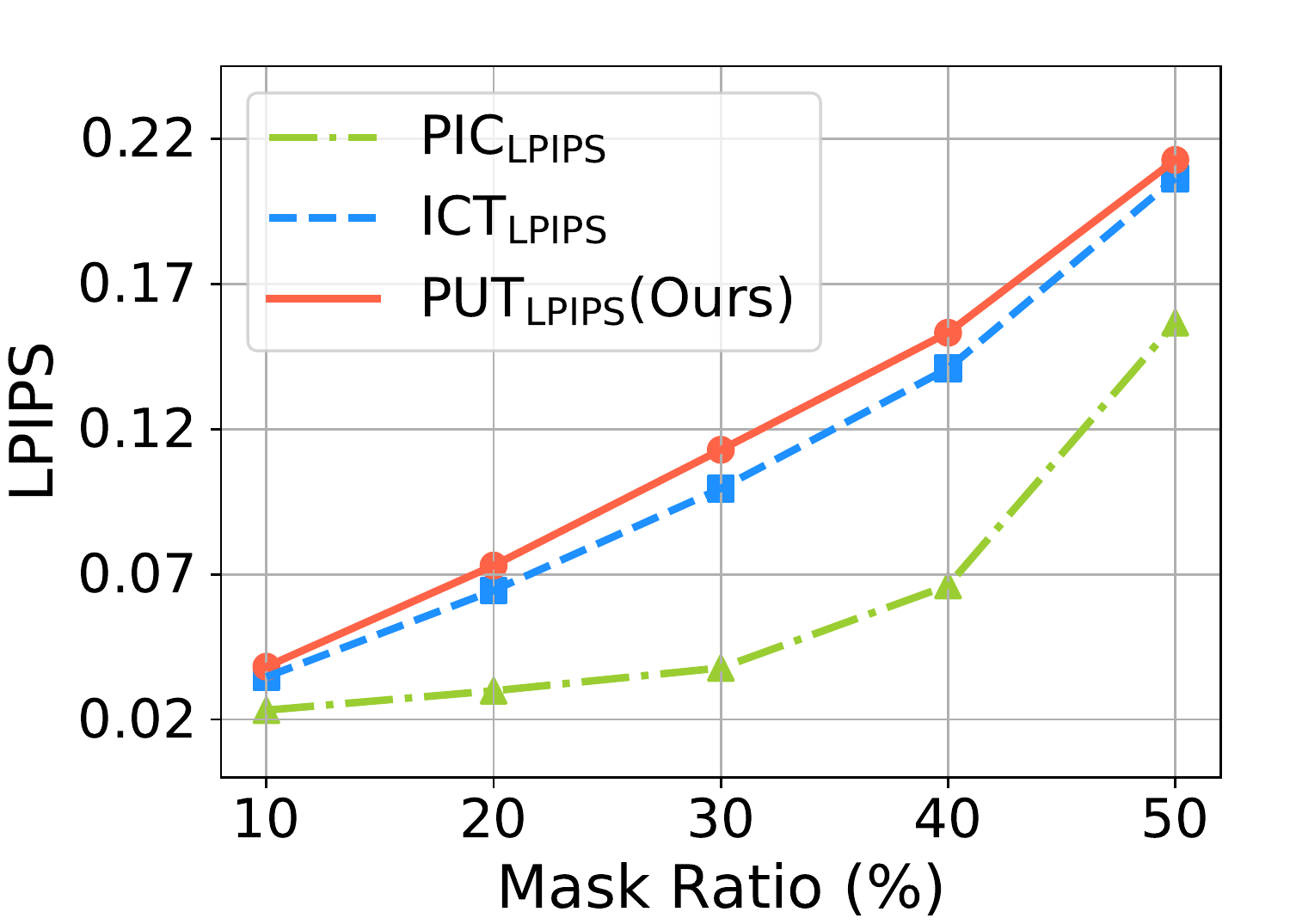}
		\end{minipage}%
	}%
	
	\subfloat[FID on ImageNet \cite{deng2009imagenet}]{
		\begin{minipage}[t]{0.5\linewidth}
			\flushleft
			\includegraphics[width=0.95\linewidth]{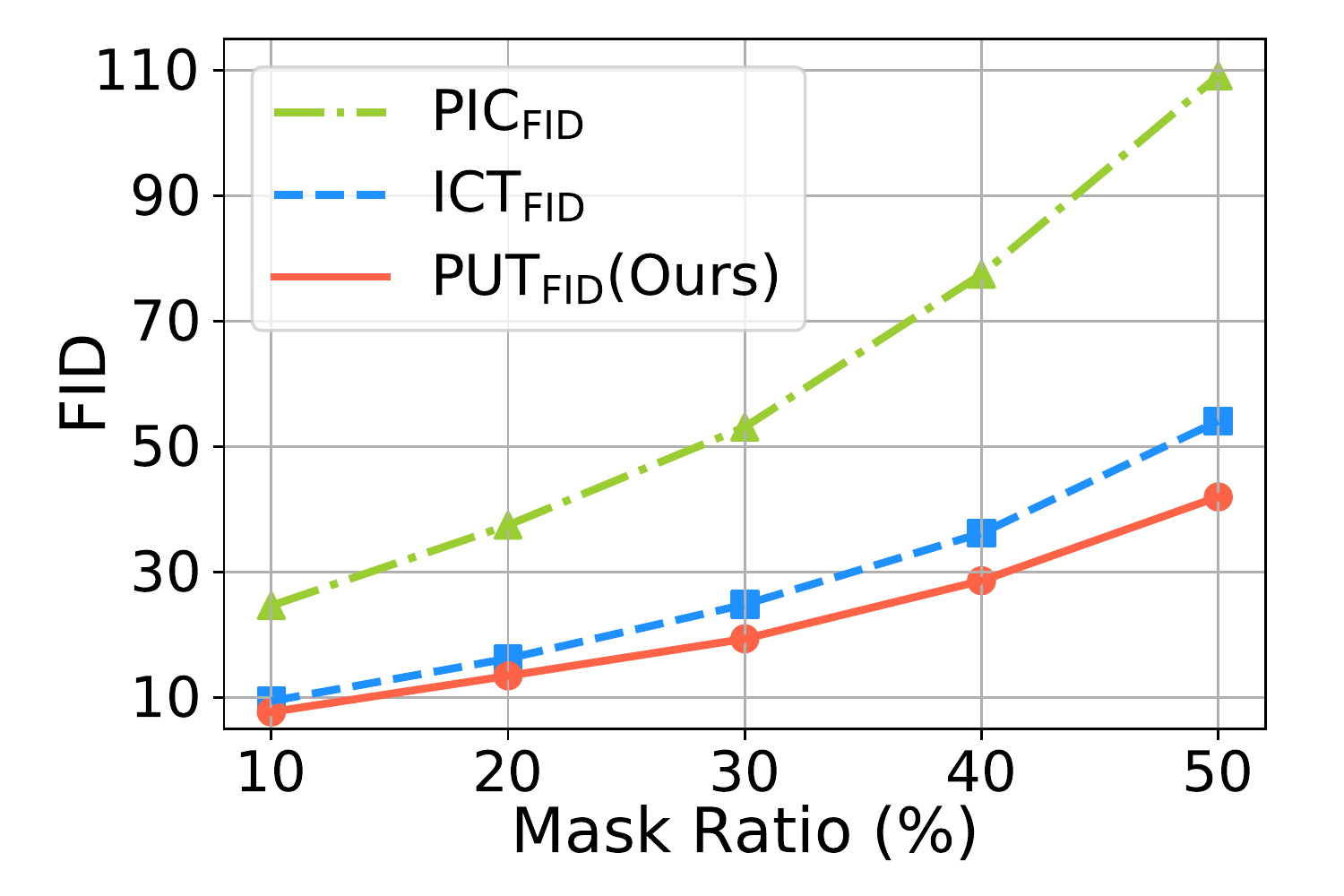}
		\end{minipage}%
	}%
	\subfloat[LPIPS on ImageNet \cite{deng2009imagenet}]{
		\begin{minipage}[t]{0.5\linewidth}
			\centering
			\includegraphics[width=0.99\linewidth]{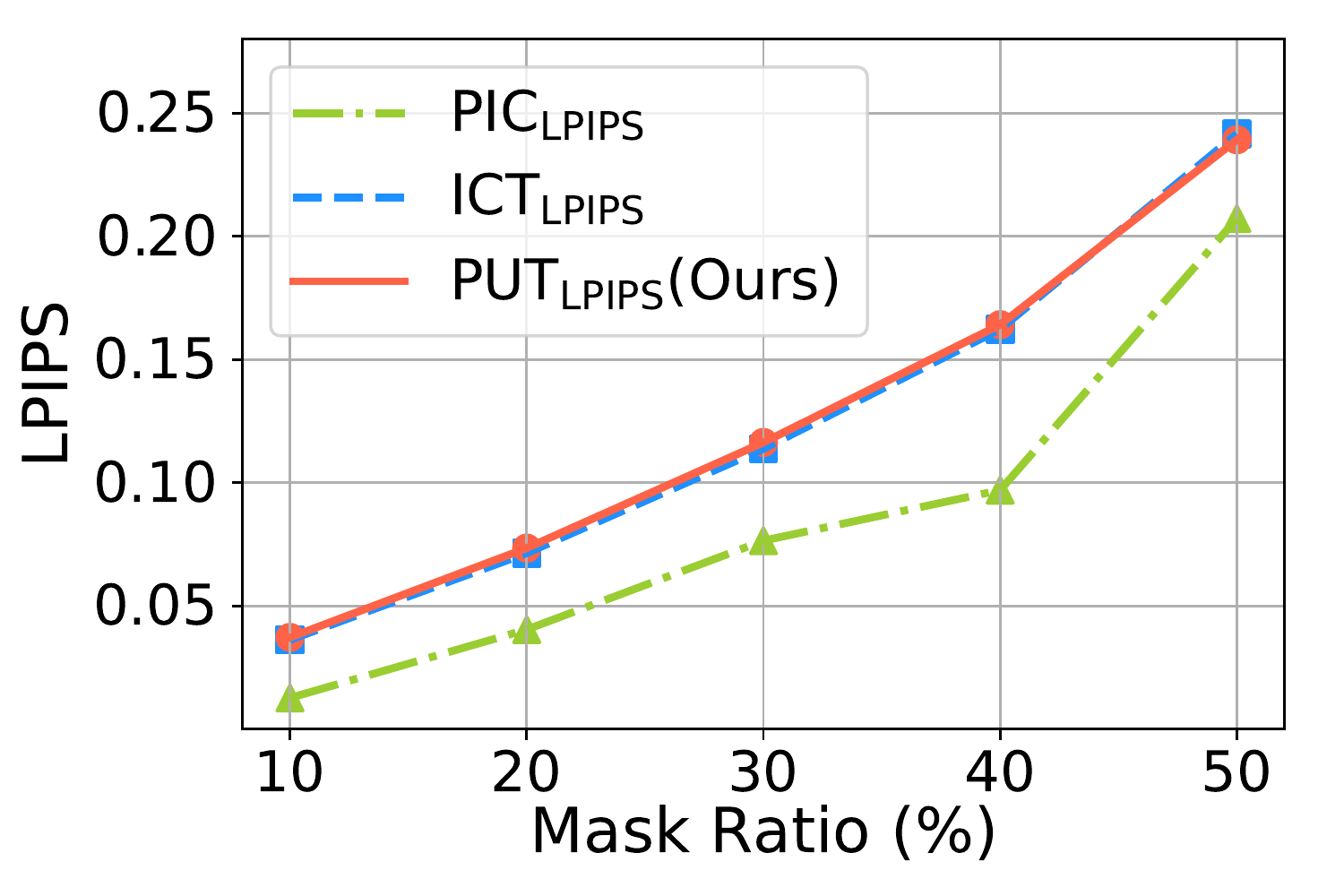}
		\end{minipage}
	}%
	\centering
	\caption{LPIPS and FID curves with respect to mask ratio on different datasets. For PUT and ICT, $\mathcal{K}=50$.}
	\vspace{-10pt}
	\label{fig: lpips_and_fid}
\end{figure}


\begin{table*}[t]
	\setlength{\tabcolsep}{8.5pt}

	\footnotesize
	\centering
		\begin{tabular}{c|c|ccc|ccc|ccc}
            \hline
            \multicolumn{2}{c|}{Dataset} & \multicolumn{3}{c|}{FFHQ \cite{karras2019style}} & \multicolumn{3}{c|}{Places2 \cite{zhou2017places}} & \multicolumn{3}{c}{ImageNet \cite{deng2009imagenet}} \\ 
            \hline
            \multicolumn{2}{c|}{Mask Ratio (\%)} & 20-40 & 40-60 & 10-60 & 20-40 &40-60 & 10-60 & 20-40 & 40-60 & 10-60 \\
            \hline
            \multirow{8}*{FID $\downarrow$} 
            & DFv2 (ICCV, 2019) \cite{yu2019free} & 27.344 & 47.894 & 30.509 & 53.107 & 83.979 & 59.280 & 49.900 & 102.111 & 64.056\\
            & EC (ICCVW, 2019) \cite{nazeri2019edgeconnect}  & 12.949 & 26.217 & 16.961 & 20.180 &34.965 & \textbf{\underline{23.206}} & 27.821 & 63.768 & 39.199\\
            & MED (ECCV, 2020) \cite{liu2020rethinking} &13.999 & 26.252 & 17.061 & 28.671 &46.815 &32.494 &40.643 &93.983 &54.854 \\
            & ICT$_{\rm all}$  (ICCV, 2021) \cite{wan2021high} &\textbf{\underline{10.442}} &23.946 &15.363 &\textbf{\underline{19.309}} &\textbf{\underline{33.510}} &23.331 &23.889 &54.327 &32.624 \\
            & PUT$_{\rm all}$ (Ours) &11.221 &\textbf{\underline{19.934}} &\textbf{\underline{13.248}} &19.776 &38.206 &24.605 & \textbf{\underline{19.411}} &\textbf{\underline{43.239}} &\textbf{\underline{26.223}}\\
            \cline{2-11}
            & PIC (CVPR, 2019) \cite{zheng2019pluralistic} & 22.847 & 37.762 & 25.902 & 31.361 & 44.289 & 34.520 & 49.215 & 102.561 & 63.955 \\
            & ICT$_{50}$ (ICCV, 2021) \cite{wan2021high} &13.536 &23.756 &16.202 &20.900 & 33.696 &24.138 &25.235 &55.598 &34.247  \\  
            & PUT$_{50}$(Ours) & \textbf{\underline{12.784}} &\textbf{\underline{21.382}} &\textbf{\underline{14.554}} &\textbf{\underline{19.617}} &\textbf{\underline{31.485}} & \textbf{\underline{22.121}} &\textbf{\underline{21.272}} &\textbf{\underline{45.153}} &\textbf{\underline{27.648}}\\
            \hline
            \multirow{8}*{PSNR $\uparrow$} 
            & DFv2 (ICCV, 2019) \cite{yu2019free} & 27.937 & 22.984 &26.783 & 26.292 & 22.412 & 25.391 & 24.464 & 20.157 & 23.387\\
            & EC (ICCVW, 2019) \cite{nazeri2019edgeconnect} & 27.484 & 22.574 & 26.181 & 26.536 &22.755 & \textbf{\underline{25.975}} & 24.703 &20.459 &23.596\\
            & MED (ECCV, 2020) \cite{liu2020rethinking} &27.117 & 22.499 & 26.111 & 25.401 &21.543 &24.510 &23.730 &19.560 &22.752\\
             & ICT$_{\rm all}$ (ICCV, 2021) \cite{wan2021high} &\textbf{\underline{29.847}} &23.041 &26.736 &25.836 &22.120 &24.986 &24.249 &20.045 &23.317 \\
             & PUT$_{\rm all}$ (Ours) &28.356 &\textbf{\underline{24.125}} &\textbf{\underline{27.473}} &\textbf{\underline{26.580}} &\textbf{\underline{22.945}} &25.749 &\textbf{\underline{25.721}} &\textbf{\underline{21.551}} &\textbf{\underline{24.726}} \\
            \cline{2-11}
            &PIC (CVPR, 2019) \cite{zheng2019pluralistic} & 25.157 & 20.424 & 24.093 & 24.073 & 20.656 & 23.469 & 22.921 & 18.368 & 21.623\\
            & ICT$_{50}$ (ICCV, 2021) \cite{wan2021high} &26.462 &21.816 &25.515 &24.947 &21.126 &24.373 &23.252 &19.025 &22.123\\
            & PUT$_{50}$(Ours) &\textbf{\underline{26.877}} &\textbf{\underline{22.375}} &\textbf{\underline{25.943}} &\textbf{\underline{25.452}} &\textbf{\underline{21.528}} &\textbf{\underline{24.492}} &\textbf{\underline{24.238}} &\textbf{\underline{19.742}} &\textbf{\underline{23.264}}\\
            \hline
            \multirow{8}*{SSIM$\uparrow$} 
            & DFv2 (ICCV, 2019) \cite{yu2019free} & 0.945 & 0.850 & 0.912 & 0.878 &0.741 & 0.831 & 0.876 &0.719 & 0.819 \\
            & EC (ICCVW, 2019) \cite{nazeri2019edgeconnect} & 0.941 & 0.826 & 0.899 &0.881 & 0.734 & \textbf{\underline{0.840}} & 0.882 & 0.714 &0.824 \\
            & MED (ECCV, 2020) \cite{liu2020rethinking} &0.936 &0.840 &0.903 &0.854 &0.685 &0.796 &0.861 &0.675 &0.795 \\
             & ICT$_{\rm all}$ (ICCV, 2021) \cite{wan2021high} &\textbf{\underline{0.964}} &0.863 &\textbf{\underline{0.917}} &0.870 &0.723 &0.819 &0.876 &0.711 &0.818\\
             & PUT$_{\rm all}$ (Ours) &0.953 &\textbf{\underline{0.888}} &0.908 &\textbf{\underline{0.885}} &\textbf{\underline{0.756}} &\textbf{\underline{0.840}} &\textbf{\underline{0.904}} &\textbf{\underline{0.772}} &\textbf{\underline{0.838}} \\
            \cline{2-11}
            &PIC (CVPR, 2019) \cite{zheng2019pluralistic} & 0.910 & 0.769 & 0.865 & 0.824 & 0.648 & 0.775 & 0.842 & 0.623 & 0.766\\
            & ICT$_{50}$ (ICCV, 2021) \cite{wan2021high} &0.931 &0.822 &0.896 &0.850 &0.682 &0.803 &0.852 &0.666 &0.786 \\
            & PUT$_{50}$(Ours) &\textbf{\underline{0.936}} &\textbf{\underline{0.845}} &\textbf{\underline{0.906}} &\textbf{\underline{0.861}} &\textbf{\underline{0.703}} &\textbf{\underline{0.806}} &\textbf{\underline{0.875}} &\textbf{\underline{0.704}} &\textbf{\underline{0.818}}\\
            \hline
            \multirow{8}*{MAE$\downarrow$} 
            & DFv2 (ICCV, 2019) \cite{yu2019free} & 0.0187 & 0.0429 & 0.0270 & 0.0230 & 0.0461 & 0.0304 & 0.0303 & 0.0638 & 0.0415 \\
            & EC (ICCVW, 2019) \cite{nazeri2019edgeconnect} & 0.0177 & 0.0430 & 0.0263 & 0.0207 &0.0419 & \textbf{\underline{0.0261}} & 0.0271 &0.0582 &0.0375 \\
            & MED (ECCV, 2020) \cite{liu2020rethinking} &0.0200 & 0.0430 &0.0277 &0.0255 &0.0505 &0.0336 &0.0320 &0.0676 &0.0434 \\
             & ICT$_{\rm all}$ (ICCV, 2021) \cite{wan2021high} &\textbf{\underline{0.0129}} &0.0368 &0.0232 &0.0221 &0.0433 &0.0289 &0.0362 &0.0578 &0.0378 \\
             & PUT$_{\rm all}$ (Ours) &0.0159 &\textbf{\underline{0.0328}} &\textbf{\underline{0.0213}} &\textbf{\underline{0.0205}} &\textbf{\underline{0.0398}} &0.0269 &\textbf{\underline{0.0233}} &\textbf{\underline{0.0487}} &\textbf{\underline{0.0321}} \\
            \cline{2-11}
            &PIC (CVPR, 2019) \cite{zheng2019pluralistic} & 0.0251 & 0.0571 & 0.0350 & 0.0284 & 0.0544 & 0.0353 & 0.0361 & 0.0785 & 0.0509\\
            & ICT$_{50}$ (ICCV, 2021) \cite{wan2021high} &0.0196 &0.0445 &0.0270 &0.0245 &0.0487 &\textbf{\underline{0.0312}} &0.0312 &0.0677 &0.0440\\
            & PUT$_{50}$(Ours) &\textbf{\underline{0.0191}} &\textbf{\underline{0.0417}} &\textbf{\underline{0.0263}} &\textbf{\underline{0.0235}} &\textbf{\underline{0.0479}} &0.0317 &\textbf{\underline{0.0281}} &\textbf{\underline{0.0641}} &\textbf{\underline{0.0401}}\\
			\hline
		\end{tabular}
	\vspace{-5pt}
	\caption{Quantitative results of different methods. 
	The methods are divided into deterministic and pluralistic groups. The subscript ``50'' of ICT and PUT is the value of $\mathcal{K}$, while the subscript ``all'' of ICT and PUT means all tokens are sampled at one iteration.} 
	\label{tab: comparison_with_other_methods_self_test}
	\vspace{-10pt}
\end{table*}

\subsection{Implementation Details}
\label{sec: implementation_details}
We use P-VQVAE with the same model size and UQ-Transformer with different model sizes for different datasets. The number of latent vectors in dual-codebook (i.e.,$K$ and $K^{'}$) both are set to 512. The detailed architecture of P-VQVAE and UQ-Transformer could be found in the supplementary material.
We train P-VQVAE with batch size 128 and train UQ-Transformer with batch size 48 for FFHQ and 96 for Places2 and ImageNet.
The learning rate is warmed up from 0 to 2e-4 and 3e-4 in the the first 5K iterations for P-VQVAE and UQ-Transformer, and then decayed with cosine scheduler.
P-VQVAE is optimized with Adam \cite{kingma2014adam} ($\beta_1=0,\beta_2=0.9$) and UQ-Transformer is optimized with AdamW \cite{loshchilov2017decoupled} ($\beta_1=0.9$, $\beta_2=0.95$). All models are trained to their convergence.

\subsection{Main Results}
\label{sec: comparison_with_other_methods}

We compare the proposed PUT with the following state-of-the-art inpainting approaches: DeepFillv2 (DFv2) \cite{yu2019free}, Edge-Connect (EC) \cite{nazeri2019edgeconnect}, MED \cite{liu2020rethinking}, PIC \cite{zheng2019pluralistic} and ICT \cite{wan2021high}. Among them, the last two ones can generate pluralistic results for each input while the others can only produce one deterministic result for the given input. For a fair comparison, we directly use the pre-trained models provided by the authors when available, otherwise train the models by ourselves using the codes and settings provided by the authors.

\paragraph{Qualitative Comparisons.} We first qualitatively compare the inpainted results of different methods in \Fref{figure: inpainting_results}. 
Specifically,  DFv2 and EC generally produce blurry images, while the texture of the results generated by MED and PIC contain lots of unnatural artifacts. Compared with ICT, PUT is more powerful in the understanding of global context and maintaining the meaningful textures, as shown in the last row in \Fref{figure: inpainting_results}.
We argue that the superiority of PUT is mainly due to: 1) high-resolution and 2) un-quantized transformer. Both of them are vital for preserving the information contained in the input images, which are helpful to produce photo-realism images.

\paragraph{User Study.} For the evaluation of subjective quality, the user study is further conducted. Only pluralistic methods are evaluated, including PIC \cite{zheng2019pluralistic}, ICT \cite{wan2021high}] and PUT. Specifically, we randomly sample 20 pairs of image and mask from the test set of each dataset. For each pair, we generate two inpainted results using each method, and ask the participants to rank these six images according to their photo-realism from high to low. 
We calculate the ratio of each method among the rank 1 images.
Results are shown in \Fref{figure: usr_study}. Our method takes up at least 60\% of the rank 1 images, demonstrating its superiority.

\paragraph{Quantitative Comparisons.} We further demonstrate the superiority of PUT in diversity and fidelity with other pluralistic methods. Specifically, the mean LPIPS distance \cite{zhang2018unreasonable}  between pairs of randomly generated results for the same input image is calculated. Following ICT \cite{wan2021high}, five pairs per input image are generated. Meanwhile, the Fréchet Inception Distance (FID) \cite{heusel2017gans} is also computed between the inpainted images and ground-truth images to reveal the fidelity of generated results. The curves of LPIPS and FID are shown in \Fref{fig: lpips_and_fid}. It can be seen that among the three pluralistic methods, PUT achieves the best fidelity (lowest FID) on all datasets, especially for large area of masked regions and complex scenes (ImageNet \cite{deng2009imagenet}). Although PUT and ICT both are implemented with transformers, 
PUT presents a higher diversity, we owe this to the feeding of original continuous feature vectors without quantization to transformer,
which has no information loss issue.

In \Tref{tab: comparison_with_other_methods_self_test}, we compare all methods in terms of several metrics, including the peak signal-to-noise (PSNR), structural similarity index (SSIM), relative L1 (MAE) and FID. 
Only one recovered output is produced for each input. Overall, for pluralistic methods, PUT$_{\rm{50}}$ performs the best in almost all metrics on all datasets. Specifically, the FID score of PUT$_{\rm 50}$ on ImageNet with mask ratio 40\%-60\% is 10.44 lower than that of ICT$_{\rm 50}$. For deterministic methods, PUT$_{\rm all}$ also achieves almost the best performance. However, the metrics are suitable for the comparison between pluralistic and deterministic methods since diverse and meaningful results can be generated by pluralistic methods.

\begin{table}[t]
	\setlength{\tabcolsep}{0.0pt}
	\footnotesize
    \centering
    \begin{tabular}{c|c|c|c|c}
        \hline
        Dataset &Metric & PUT$_{\rm all}$ (Ours) &CoModGAN \cite{zhao2021comodgan}
        & LaMa\cite{suvorov2022resolution} \\
        \hline
        FFHQ \cite{karras2019style} &PSNR $\uparrow$ / FID $\downarrow$ &\underline{\bf 24.245}/21.351 &22.430/\underline{\bf 17.914} &-/- \\
        \hline
        Places2 \cite{zhou2017places} & PSNR $\uparrow$ / FID $\downarrow$ &22.589/38.472 &20.962/32.559 &\underline{\bf 22.694}/\underline{\bf 31.160}   \\
        \hline
    \end{tabular}
    \caption{Performance of different methods on FFHQ \cite{karras2019style} and Places2 \cite{zhou2017places} with mask ratio 40\%-60\% and resolution $512\times 512$.}
    \label{tab:comparison_sota_ffhq_places2}
\end{table}

In Table \ref{tab:comparison_sota_ffhq_places2}, we further compare PUT$_{\rm all}$ with LaMa \cite{suvorov2022resolution} and CoModGAN \cite{zhao2021comodgan}, both of which are recently proposed methods for image inpainting with resolution $512 \times 512$. For PUT$_{\rm all}$, an upsample network used in ICT \cite{wan2021high} is trained to upsample the results from $256\times256$ to $512\times512$. As we can see, PUT$_{\rm all}$ achieves a much better PSNR than CoModGAN, and it is also comparable with LaMa.

\begin{table}[t]
	\setlength{\tabcolsep}{1.5pt}
	\footnotesize
	\centering
		\begin{tabular}{c|c|ccc|ccc}
            \hline
            \multicolumn{2}{c|}{Dataset} & \multicolumn{3}{c|}{FFHQ \cite{karras2019style}} & \multicolumn{3}{c}{Places2 \cite{zhou2017places}} \\ 
            \hline
            \multicolumn{2}{c|}{Mask Ratio (\%)} & 20-40 & 40-60 & 10-60 & 20-40 &40-60 & 10-60 \\
            \hline
            \multirow{6}*{\rotatebox{0}{FID$\downarrow$}} 
            & PUT$^{\rm conv}$ &163.610 &226.437 &173.351 &178.057 &216.235 &179.294  \\ 
            & PUT$^{\rm one}$ &\textbf{\underline{12.112}} &\textbf{\underline{20.298}} &\textbf{\underline{13.960}} &26.015 &38.011 &28.634 \\ 
            & PUT$^{\rm no\_ref}$ &15.014 &22.736 &16.469 &22.821 &32.281 &25.084 \\
            & PUT$^{\rm tok}$ &22.824 &38.384 &26.098 &58.643 &120.482 &75.625  \\ 
            & PUT$^{\rm qua0}$ &43.621 &96.648 & 54.879 &35.127 &71.629 &44.588 \\ 
            & PUT &12.784 &21.382 &14.554 &\textbf{\underline{19.617}} &\textbf{\underline{31.485}} &\textbf{\underline{22.121}}  \\ 
            \hline
            \multirow{6}*{\rotatebox{0}{PSNR$\uparrow$}} 
            & PUT$^{\rm conv}$ &12.783 &9.735 &12.360 &12.207 &9.347 &11.799  \\
            & PUT$^{\rm one}$ &\textbf{\underline{26.887}} &22.335 &25.903 &24.507 &20.571 &23.600  \\ 
            & PUT$^{\rm no\_ref}$ &26.487 &22.249 &25.547 &25.053 &21.398 &24.185 \\
            & PUT$^{\rm tok}$ &23.916 &18.936 &22.879 &20.940 &14.685 &19.429 \\ 
            & PUT$^{\rm qua0}$ &24.188 &19.564 &23.174 &24.340 &20.239 &23.353 \\ 
            & PUT  &26.877 &\textbf{\underline{22.375}} &\textbf{\underline{25.943}} &\textbf{\underline{25.452}} &\textbf{\underline{21.528}} &\textbf{\underline{24.492}} \\ 
            \hline
            \multirow{6}*{\rotatebox{0}{SSIM$\uparrow$}} 
            & PUT$^{\rm conv}$ &0.495 &0.266 &0.445 &0.417 &0.212 &0.373  \\  
            & PUT$^{\rm one}$  &0.937 &0.844 &0.906 &0.836 &0.660 &0.776 \\
            & PUT$^{\rm no\_ref}$ &0.932 &0.841 &0.901 &0.851 &0.695 &0.798 \\
            & PUT$^{\rm tok}$ &0.893 &0.745 &0.843 &0.742 &0.453 &0.652 \\ 
            & PUT$^{\rm qua0}$ &0.879 &0.704 &0.820 &0.831 &0.637 &0.763 \\ 
            & PUT &\textbf{\underline{0.945}} &\textbf{\underline{0.857}} &\textbf{\underline{0.914}} & \textbf{\underline{0.861}} &\textbf{\underline{0.703}} &\textbf{\underline{0.806}}   \\ 
            \hline
            \multirow{6}*{\rotatebox{0}{MAE$\downarrow$}} 
            & PUT$^{\rm conv}$ &0.1420 &0.3017 &0.1912 &0.1447 &0.2995 &0.1941 \\ 
            & PUT${\rm one}$ &\textbf{\underline{0.0191}} &0.0418 &0.0264 &0.0256 &0.0526 &0.0345  \\ 
            & PUT$^{\rm no\_ref}$ &0.0202 &0.0426 &0.0274 &0.0249 &0.0488 &0.0329\\
            & PUT$^{\rm tok}$ &0.0289 &0.0689 &0.0418 &0.0467 &0.1337 &0.0779 \\ 
            & PUT$^{\rm qua0}$ &0.0281 &0.0627 &0.0393 &0.0269 &0.0562 &0.0367 \\ 
            & PUT &\textbf{\underline{0.0191}} &\textbf{\underline{0.0417}} &\textbf{\underline{0.0263}} &\textbf{\underline{0.0235}} &\textbf{\underline{0.0479}} &\textbf{\underline{0.0317}} \\ 
			\hline
		\end{tabular}
	\caption{Quantitative results of different methods. All the results are tested with $\mathcal{K}=50$.} 
	\vspace{-8pt}
	\label{tab: ablation_study_results}
\end{table}

\subsection{Discussions}
\label{sec: ablation_study} 
\begin{figure}[t]
	\centering
	\includegraphics[width=1.0\columnwidth]{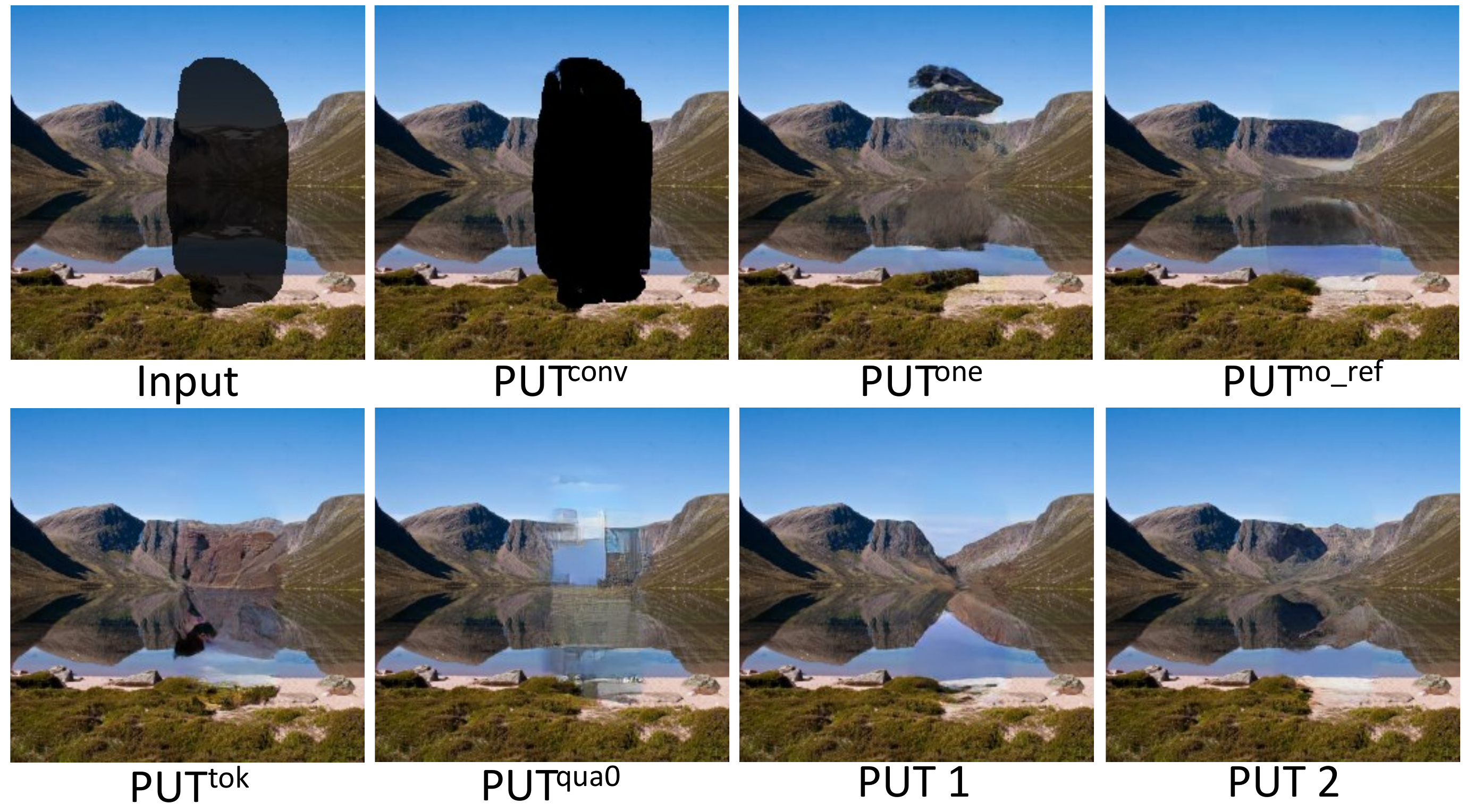} 
	\vspace{-15pt}
	\caption{Inpainted results of different models on Places2 \cite{zhou2017places}.}
	\vspace{-10pt}
	\label{figure: ablation_inpainting_results}
\end{figure}

\paragraph{Effectiveness of different components.}
To show the effectinesess of the patch-based encoder, dual-codebook, multi-scale guided decoder and un-quantized transformer,
several methods are designed: 1) PUT$^{\rm conv}$ means that the patch-based encoder is replaced with a normal CNN encoder, which is implemented with convolution layers; 2) PUT$^{\rm one}$ refers that there is only one codebook for vector quantization; 3) PUT$^{{\rm no\_ref}}$ means that there is no reference branch in the decoder; and 4) PUT$^{\rm tok}$ denotes that the transformer takes the quantized vectors as input rather than the original feature vectors from encoder.
In order to show the effectiveness of random quantization of feature vectors while training UQ-Transformer, we further design the fifth model, PUT$^{\rm qua0}$, that trains UQ-Transformer without random quantization. For all models, except the modifications mentioned, others remain the same with our default model PUT. Results are shown in \Tref{tab: ablation_study_results} and \Fref{figure: ablation_inpainting_results}.

Among all models, PUT$^{\rm conv}$ performs the worst in all metrics, demonstrating the effectiveness of the non-overlapping patch partition design. Within CNN based encoder, the input images are processed in a sliding window manner, introducing the interaction between masked and unmasked regions, which is fatal to transformer for the prediction of masked regions. 

Comparing PUT$^{\rm one}$ to PUT, the only difference is the training of P-VQVAE with one or two codebooks 
since the codebook $\mathbf{e'}$ is not used in the inference stage. 
P-VQVAE can learn more discriminative features for masked and unmasked patches with the help of dual-codebook. Interestingly, PUT indeed performs better than PUT$^{\rm one}$ except the FID score on FFHQ. We speculate that face generation is much easier because all faces share a similar structure. Besides, the facial structure is related with the position in the image since most of the images in FFHQ contain faces in their near-center locations. As we can see in \Fref{figure: ablation_inpainting_results}, PUT$^{\rm one}$ sometimes predicts \emph{black} patches, which is similar to those patches containing missing pixels. Nonetheless, PUT achieves overall better performance than PUT$^{\rm one}$.

Compared with PUT, PUT$^{\rm no\_ref}$ constructs the inpainted image without referencing to the input masked image, which leads to a inferior performance in terms of all metrics. For PUT, some useful textures can be recovered with the help of the guidance from the unmasked regions in reference image. As we can see in \Fref{figure: ablation_inpainting_results}, the result of PUT$^{\rm no\_ref}$ is over smoothed, which is unnatural.

Compared with PUT$^{\rm tok}$, PUT performs much better in all metrics. Without quantizing feature vectors to discrete representations, no information contained in the feature will be lost. Such practice helps transformer to understand complex content and maintain the inherit meaningful textures in the input image. However, the training of UQ-Transformer should be carefully designed by randomly quantizing the input feature vectors since only quantized vectors can be obtained for masked regions at the inference stage. The effectiveness of such random quantization during training is obvious while comparing PUT with PUT$^{\rm qua0}$.

\begin{figure}[t]
	\centering
	\includegraphics[width=1.0\columnwidth]{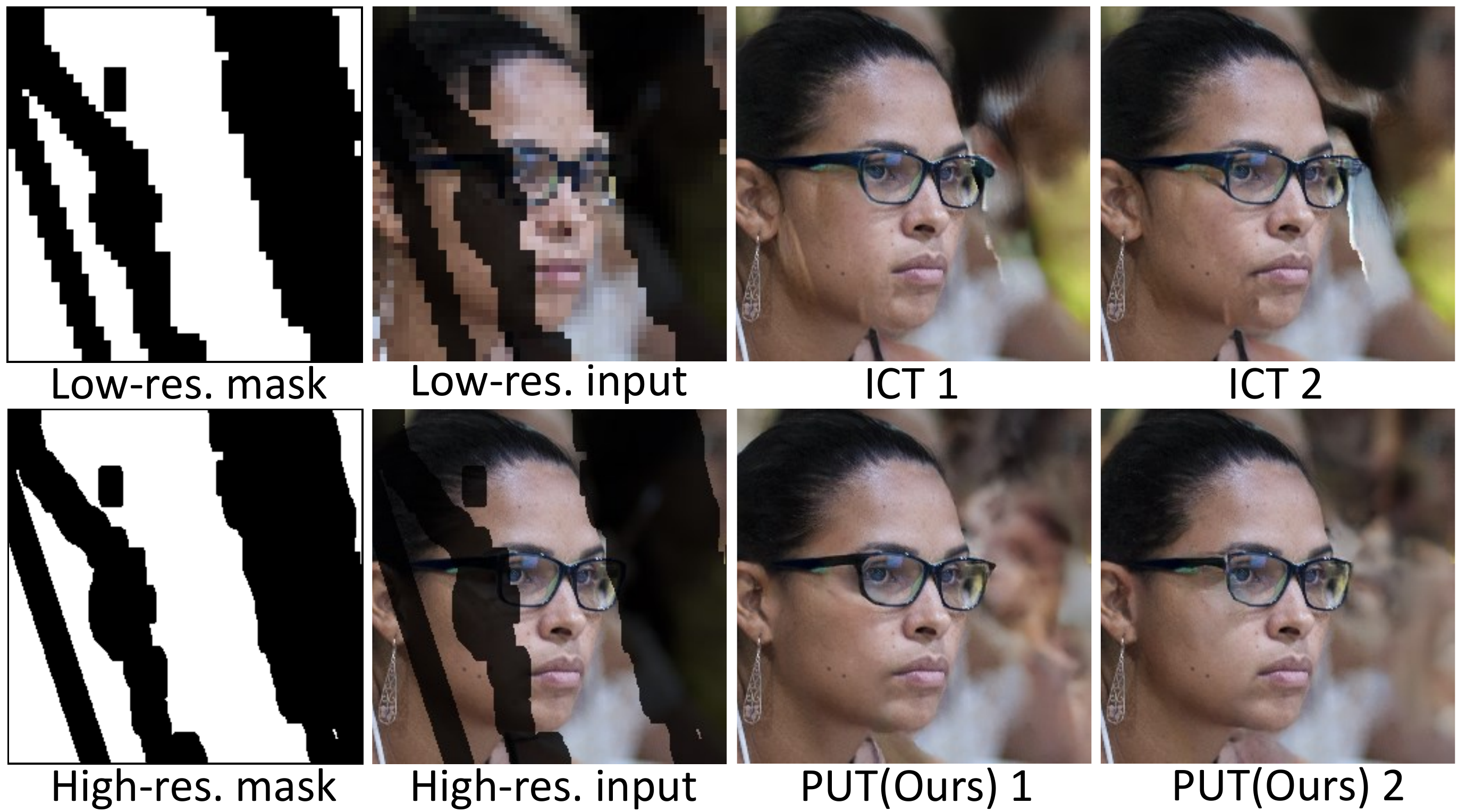} 
	\vspace{-15pt}
	\caption{Misalignment between the high and low resolutions. The mask and input image in low-resolution are zoomed in for a better comparison.}
	\vspace{-10pt}
	\label{figure: misalignment_between_high_and_low_resolution}
\end{figure}

\paragraph{Misalignment between high and low resolutions.} Existing transformer based methods \cite{wan2021high} involves high and low resolutions. 
The images and masks in high and low resolutions are misaligned, especially the borders of masked regions. For real applications, where the masked regions are often with arbitrary shapes and sizes, such misalignment is non-negligible. In \Fref{figure: misalignment_between_high_and_low_resolution}, we show one case that with irregular masked regions. We can see that the results of ICT \cite{wan2021high} contain lots of artifacts near the misaligned borders. For PUT, the generated results are natural and smooth even though the provided mask is irregular.

\section{Conclusions and Limitations}
\label{sec: conclusion} 
In this paper, we present a novel method, PUT, for pluralistic image inpainting. PUT consists of two main components: 1) patch-based auto-encoder (P-VQVAE) and 2) un-quantized transformer (UQ-Transformer). With the help of P-VQVAE and UQ-Transformer, PUT processes the original high-resolution image without quantization. Such practice preserves the information contained in the input image as much as possible. 
The main limitation of PUT is the inference speed for the production of diverse results. However, it is a common issue of existing transformer based auto-regressive methods \cite{vaswani2017attention, wan2021high, ramesh2021zero, esser2021taming}. Two solutions could be adopted to alleviate this limitation: 1) replacing the used transformer block with more efficient ones \cite{ho2019axial, wang2020linformer} and 2) sampling tokens for several patches at each iteration. 
In addition, PUT may be used for editing the contents of images to achieve illicit goals, which can be mitigated using existing synthesized image detectors \cite{wang2020cnn}.
Experimental results demonstrate the superiority of PUT, including the fidelity and diversity, especially for large masked regions and complex scenes (such as ImageNet \cite{deng2009imagenet}).

\vspace{12pt}
\noindent\textbf{Acknowledgement.} 
This work is supported by the National Natural Science Foundation of China (No. 62002336, No. U20B2047) and the Exploration Fund Project of University of Science and Technology of China (YD3480002001).

{\small
\bibliographystyle{ieee_fullname}
\bibliography{egbib}
}

\clearpage

\appendix

\twocolumn[{%
\renewcommand\twocolumn[1][]{#1}%
\maketitle
\begin{center}
\LARGE
\bf{Supplementary}
\vspace{30pt}
\end{center}%
}]

\section{Overview}
In this supplementary material, we provide more implementation details, experimental results and analysis, including:
\begin{itemize}
    \item training of P-VQVAE (\Sref{sec: traning_of_p_vqvae}).
    \item sampling strategy for image inpainting ( \Sref{sec: sampling_strategy_supp}).
    \item network architecture of different models (\Sref{sec: network_architecture}).
    \item more results on different datasets (\Sref{sec: more_results}). 
    
    \item more discussions on PUT (\Sref{sec: discussion}), including 
    the inference speed of PUT and some artifacts in inpainted results.
    
\end{itemize}

\section{Training of P-VQVAE}
\label{sec: traning_of_p_vqvae}
Given an image $\mathbf{x}$ and two different masks $\mathbf{m}$ and $\mathbf{m'}$, the input of P-VQVAE is $\mathbf{\hat{x}}= \mathbf{x} \otimes \mathbf{m}$. The overall loss for the training of P-VQVAE is:
\begin{equation}
\begin{aligned}
     L_{vae}= \mathcal{L}_{rec}(\mathbf{\hat{x}}, \mathbf{\hat{x}}^R) 
	 + \parallel {\rm sg}[\mathbf{\hat{f}}] \ominus \mathbf{\hat{e}} \parallel^2_2 
	 + \beta \parallel {\rm sg}[\mathbf{\hat{e}}] \ominus \mathbf{\hat{f}} \parallel^2_2,
\end{aligned}
\label{eq: p_vqvae_loss}
\end{equation}
where  $\mathbf{\hat{f}}=\mathcal{E}(\mathbf{\hat{x}})$ denotes the feature vectors extracted by the encoder and $\mathbf{\hat{e}}$ is quantized vectors for $\mathbf{\hat{f}}$.  $\mathbf{\hat{x}}^R = \mathcal{D}(\mathbf{\hat{e}}, \mathbf{m} \otimes \mathbf{m'}, \mathbf{\hat{x}}\otimes \mathbf{m'})$ is the reconstructed image and ${\rm sg}[\cdot]$ refers to a stop-gradient operation that blocks gradients from flowing into its argument.

\begin{algorithm}[tp]
	\footnotesize
	\caption{Sampling Strategy for Pluralistic Image Inpainting}
	\label{algorithm_sampling_strategy_for_image_inpainting}
	\SetKwInOut{Input}{Input}
	\SetKwInOut{Output}{Output}
	\Input{
    	$\mathbf{\hat{x}} \in \mathbb{R}^{H \times W \times 3}$: masked image needs to be inpainted \newline
    	$\mathbf{m} \in \{0,1\}^{H \times W \times 3}$: the mask indicating whether a pixel is masked/missing or not \\
    	$\mathcal{K}$: top-$\mathcal{K}$ for Gibbs sampling
	}
	\Output{
	    $\mathbf{\hat{x}}^I \in \mathbb{R}^{H \times W \times 3}$: the inpainted image
	}
	\textbf{Step1}: get indicator mask, feature vectors, quantized tokens \\
    \quad $\mathbf{m}^{\downarrow} \in \{0,1\}^{\frac{H}{r}\times \frac{W}{r} \times 1}$: calculated from $\mathbf{m}$  \\
    \quad $\mathbf{\hat{f}} \in \mathbb{R}^{\frac{H}{r} \times \frac{W}{r} \times C} \leftarrow \mathcal{E}(\mathbf{\hat{x}})$ \\
    \quad $\mathbf{\hat{t}} \in \mathbb{N} ^{\frac{H}{r} \times \frac{W}{r}} \leftarrow \mathcal{I}(\mathbf{\hat{f}}, \mathbf{e}, \mathbf{e'}, \mathbf{m}^{\downarrow})$ // Sec. 3.1 in the paper\\
    $\quad \mathbf{\hat{t}}^{I} \leftarrow \mathbf{\hat{t}}$\\
    
	\textbf{Step2}: sample tokens for masked patches \\
	\quad \While{$\sum_{i,j}\mathbf{m}^{\downarrow}_{i,j} < \frac{HW}{r^2}$}
	{
	    $\mathbf{\hat{p}} \in [0,1]^{\frac{H}{r}\times \frac{W}{r}\times K} \leftarrow \mathcal{T}(\mathbf{\hat{f}})$ // probabilities, Sec. 3.2 in the paper\\ 
	    // select the patch with maximum probability \\
	    $i',j' \leftarrow {\rm argmax}_{i,j} (1-\mathbf{m}^{\downarrow}_{i,j}) \cdot \max \mathbf{\hat{p}}_{i,j,:}$ \\ 

	    // sample the token from the top-$\mathcal{K}$ elements in $\mathbf{\hat{p}}_{i',j',:}$ \\
	    $k \leftarrow \Call{GibbsSampling}{\mathbf{\hat{p}}_{i',j',:}, \mathcal{K}}$ \\
	    
	    // update some variables  \\
	    $\mathbf{\hat{t}}^{I}_{i',j'} \leftarrow k$,
	    $\mathbf{m}^{\downarrow}_{i',j'} \leftarrow 1$, 
	    $\mathbf{\hat{f}}_{i', j'} \leftarrow \mathbf{e}_k$ \\
	}
	\textbf{Step3}: reconstruct the image\\
    \quad $\mathbf{\hat{e}}^{I} \in \leftarrow \Call{VectorRetrieval}{\mathbf{\hat{t}}^{I}, \mathbf{e}}$\\
    \quad $\mathbf{\hat{x}}^{I} \leftarrow \mathcal{D}(\mathbf{\hat{e}}^{I}, \mathbf{m}, \mathbf{\hat{x}})$ \\
	Return $\mathbf{\hat{x}}^I$
\end{algorithm}

The last term in Eq. (\ref{eq: p_vqvae_loss}) is the so-called \emph{commitment loss} \cite{van2017neural} with weighting factor $\beta = 0.25$. It is responsible for passing gradient information from decoder to encoder.
The second term in Eq. (\ref{eq: p_vqvae_loss}) is the codebook loss for the optimization of latent vectors. Following previous works in \cite{van2017neural,razavi2019generating}, we replace the second term with the Exponential Moving Average (EMA) to optimize $\mathbf{e}$ and $\mathbf{e'}$. Specifically, at each iteration $t$, the latent vector $\mathbf{e}_{k}$ is updated as:
\begin{equation}
	\begin{cases}
		n_{k}^{t} = n_{k}^{t-1} * \gamma + n_{k} * (1-\gamma),\\
		\mathbf{\Bar{e}}_{k}^{t} = \mathbf{\Bar{e}}_{k}^{t-1} * \gamma + \sum\limits_{j}^{n_{k}}(\mathbf{\hat{f}}^{k})_{j} * (1-\gamma), \\
		\mathbf{e}^{t}_{k} = \frac{\mathbf{\Bar{e}}_{k}^{t}}{n_{k}^{t}},
	\end{cases}
	\label{eq: ema_for_latent_embedding}
\end{equation}
where $\mathbf{\hat{f}}^{k}$ denotes the set of feature vectors in $\mathbf{\hat{f}}$ that assigned to $\mathbf{e}_{k}$ and $n_{k}$ is the number of feature vectors in $\mathbf{\hat{f}}^{k}$. $\gamma$ is the decay parameter with the value between 0 and 1. We set $\gamma=0.99$ in all our experiments.

The first term in Eq. (\ref{eq: p_vqvae_loss}) is the reconstruction loss and $\mathcal{L}_{rec}(\cdot, \cdot)$ is the function to get the difference between the inputted and reconstructed images. It consists of five parts, including L1 loss between the pixel values in two images (denoted as $\mathcal{L}_{pixel}$) and the gradients of two images (denoted as $\mathcal{L}_{grad}$), the adversarial loss \cite{goodfellow2014generative} $\mathcal{L}_{adv}$, as well as the perceptual loss \cite{johnson2016perceptual} $\mathcal{L}_{perc}$ and style loss\cite{gatys2016image} $\mathcal{L}_{style}$ between the two images. The design of the last three losses are inspired by the work in \cite{nazeri2019edgeconnect}. In the following, we describe the aforementioned losses in detail. Among them:
\begin{equation}
    \mathcal{L}_{pixel} = \mathcal{M}(|\mathbf{\hat{x}} \ominus \mathbf{\hat{x}}^R|),
	\label{eq: loss_pixel}
\end{equation}
\begin{equation}
    \mathcal{L}_{grad} = \mathcal{M}(|{\rm grad}[\mathbf{\hat{x}}] \ominus {\rm grad}[\mathbf{\hat{x}}^R]|),
	\label{eq: loss_gradient}
\end{equation}
where $\mathcal{M}(\cdot)$ refers to a mean-value operation, ${\rm grad}[\cdot]$ is the function calculating the gradient of the given image.

\begin{table}[t]
	\setlength{\tabcolsep}{0.5pt}
	\footnotesize
	\centering
		\begin{tabular}{c|c|c|c}
			\hline
			Module  & Layer & Parameter size / Stride & Output size \\
			\hline 
			\multirow{4}*{P-Enc}
			& Linear & $192 \times 256$ & $32 \times 32 \times 256$\\
			\cline{2-4}
			& \makecell[c]{Linear\\ResBlock} & $\left(\begin{array}{c} 256 \times 128\\ 128 \times 256\\ \end{array} \right) \times 8$    & $32 \times 32 \times 256$\\
			\cline{2-4}
			& Linear & $256 \times 256$ &$32\times32\times256$ \\
			\hline
			\multirow{2}*{D-Codes}
			& $\mathbf{e}$ & $512 \times 256$ & - \\
			\cline{2-4}
			& $\mathbf{e'}$ & $512 \times 256$ & - \\
			\hline
			\multirow{11}*{MSG-Dec}
			& Conv & $256 \times 3\times3 \times 256 / 1$ & $32 \times 32 \times 256$\\
			\cline{2-4}
			& \makecell[c]{Conv\\ResBlock} & $\left(\begin{array}{c} 256 \times 3 \times 3 \times 128/1\\ 128 \times 3\times 3 \times 256/1\\ \end{array} \right) \times 8$    & $32 \times 32 \times 256$\\
			\cline{2-4}
			& \makecell[c]{Deconv\\(Conv)} & \makecell[c]{$256 \times 4 \times 4 \times 256/2$\\($256\times4\times4\times256/2$)} &\makecell[c]{$64\times64\times256$\\($32\times32\times256$)} \\
			\cline{2-4}
			& \makecell[c]{Deconv\\(Conv)} & \makecell[c]{$256 \times 4 \times 4 \times 128/2$\\($128\times4\times4\times256/2$)} &\makecell[c]{$128\times128\times128$\\($64\times64\times256$)} \\
			\cline{2-4}
			& \makecell[c]{Deconv\\(Conv)} & \makecell[c]{$128 \times 4 \times 4 \times 64/2$\\($64\times4\times4\times128/2$)} &\makecell[c]{$256\times256\times64$\\($128\times128\times128$)} \\
			\cline{2-4}
			& \makecell[c]{Conv$^\dag$\\(Conv)} & \makecell[c]{$64 \times 3 \times 3 \times 3/1$\\($3\times3\times3\times64/1$)} &\makecell[c]{$256\times256\times3$\\($256\times256\times64$)} \\	
			\hline
		\end{tabular}
	\caption{Architecture of P-VQVAE. For MSG-Dec, the bracketed layers in the bottom four rows denotes the layers in reference branch. Except the convolution layer marked by $\dag$, all the other layers are followed by a ReLU \cite{nair2010rectified} activation function. The structure of Linear and Conv ResBlocks are shown in \Fref{figure: blocks_architecture}.}
	\label{tab: architecture_of_p_vqvae}
\end{table}

\begin{table}[t]
	\setlength{\tabcolsep}{0.5pt}
	\footnotesize
	\centering
		\begin{tabular}{c|c|c|c}
			\hline
			Module  & Layer & Parameter size / Stride & Output size \\
			\hline 
			\multirow{6}*{Conv-Enc}
		    & Conv & $3 \times 4 \times 4 \times 64 / 2$ & $ 128 \times 128 \times 64$\\
			\cline{2-4}
			& Conv & $64 \times 4 \times 4 \times 128 / 2$ & $ 64 \times 64 \times 128$\\
			\cline{2-4}
			& Conv & $128 \times 4 \times 4 \times 256 / 2$ & $ 32 \times 32 \times 256$\\
			\cline{2-4}
			& \makecell[c]{Conv\\ResBlock} & $\left(\begin{array}{c} 256 \times 3 \times 3 \times 128/1\\ 128 \times 3\times 3 \times 256/1\\ \end{array} \right) \times 8$    & $32 \times 32 \times 256$\\
			\cline{2-4}
			& Conv & $256 \times 3 \times 3 \times 256$ &$32\times32\times256$ \\
			\hline
		\end{tabular}
	\caption{Architecture of the encoder in P-VQVAE$^{\rm conv}$. The learnable codebook and decoder are the same with those in P-VQVAE in \Tref{tab: architecture_of_p_vqvae}. All layers are followed by a ReLU \cite{nair2010rectified} activation function.}
	\label{tab: architecture_of_p_vqvae_conv}
\end{table}

The adversarial loss $\mathcal{L}_{adv}$ is computed with the help of a discriminator network $\mathcal{D}_{adv}(\cdot)$:
\begin{equation}
    \mathcal{L}_{adv} = - \mathcal{M}({\rm log}[1 \ominus \mathcal{D}_{adv}(\mathbf{\hat{x}}^R)]) - \mathcal{M}({\rm log}[\mathcal{D}_{adv}(\mathbf{\hat{x}})]),
	\label{eq: loss_adv}
\end{equation}
where ${\rm log}[\cdot]$ denotes element-wise logarithm operation. 
The architecture of the discriminator network is the same with that in \cite{nazeri2019edgeconnect}.

The conceptual loss $\mathcal{L}_{perc}$ and style loss $\mathcal{L}_{style}$ are computed based on the activation maps from VGG-19 \cite{simonyan2014very}:
\begin{equation}
    \mathcal{L}_{perc} = \sum\limits_{l}^{L_{perc}} \mathcal{M}(|\phi_{l}(\mathbf{\hat{x}}) \ominus \phi_{l}(\mathbf{\hat{x}}^R)|)
	\label{eq: loss_perc}
\end{equation}
\begin{equation}
    \mathcal{L}_{style} = \sum\limits_{l}^{L_{style}} \mathcal{M}(|\mathcal{G}(\phi_{l}(\mathbf{\hat{x}})) \ominus \mathcal{G}(\phi_{l}(\mathbf{\hat{x}}^R))|)
	\label{eq: loss_style}
\end{equation}
where $\phi_{l}(\cdot)$ corresponds to different layers in VGG-19 \cite{simonyan2014very}, $\mathcal{G}(\cdot)$ denotes the function that gets the Gram matrix of its argument. For $\mathcal{L}_{perc}$ and $\mathcal{L}_{style}$, we set $L_{perc}=\{{\rm relu1\_1}, {\rm relu2\_1}, {\rm relu3\_1}, {\rm relu4\_1}, {\rm relu5\_1}\}$ and  $L_{perc}=\{{\rm relu2\_2}, {\rm relu3\_4}, {\rm relu4\_4}, {\rm relu5\_2}\}$. The overall reconstruction loss is:
\begin{equation}
\begin{aligned}
    \mathcal{L}_{rec} & = \mathcal{L}_{pixel} + \lambda_{g} \mathcal{L}_{grad} + \lambda_{a} \mathcal{L}_{adv} \\
    & + \lambda_{p} \mathcal{L}_{perc} + \lambda_{s} \mathcal{L}_{style}
\end{aligned}
	\label{eq: loss_reconstruction}
\end{equation}
In our implementation, we set $\lambda_{g}=5$,  $\lambda_{a}=0.1$,  $\lambda_{p}=0.1$ and  $\lambda_{s}=250$.

\section{Sampling Strategy for Image Inpainting}
\label{sec: sampling_strategy_supp}
 The overall procedure can be divided into three steps: 1) get the feature vectors $\mathbf{\hat{f}}$ from the masked image $\mathbf{\hat{x}}$ using encoder and get the tokens $\mathbf{\hat{t}}$ by quantizing $\mathbf{\hat{f}}$ with latent vectors in dual-codebook. The tokens for masked patches are not required; 2) get the tokens for masked patches using transformer. Note that the tokens are iteratively sampled with Gibbs sampling following previous transformer-based works \cite{radford2019language, ramesh2021zero, esser2021taming}; 3) retrieve quantized vectors $\mathbf{\hat{e}}^{I}$ from codebook $\mathbf{e}$ based on the tokens and reconstruct the inpainted image $\mathbf{\hat{x}}^{I}$ using decoder by referencing to masked image $\mathbf{\hat{x}}$. The detailed sampling strategy is shown in Algorithm \ref{algorithm_sampling_strategy_for_image_inpainting}.

\begin{figure}[tp]
	\centering
	\includegraphics[width=1.0\columnwidth]{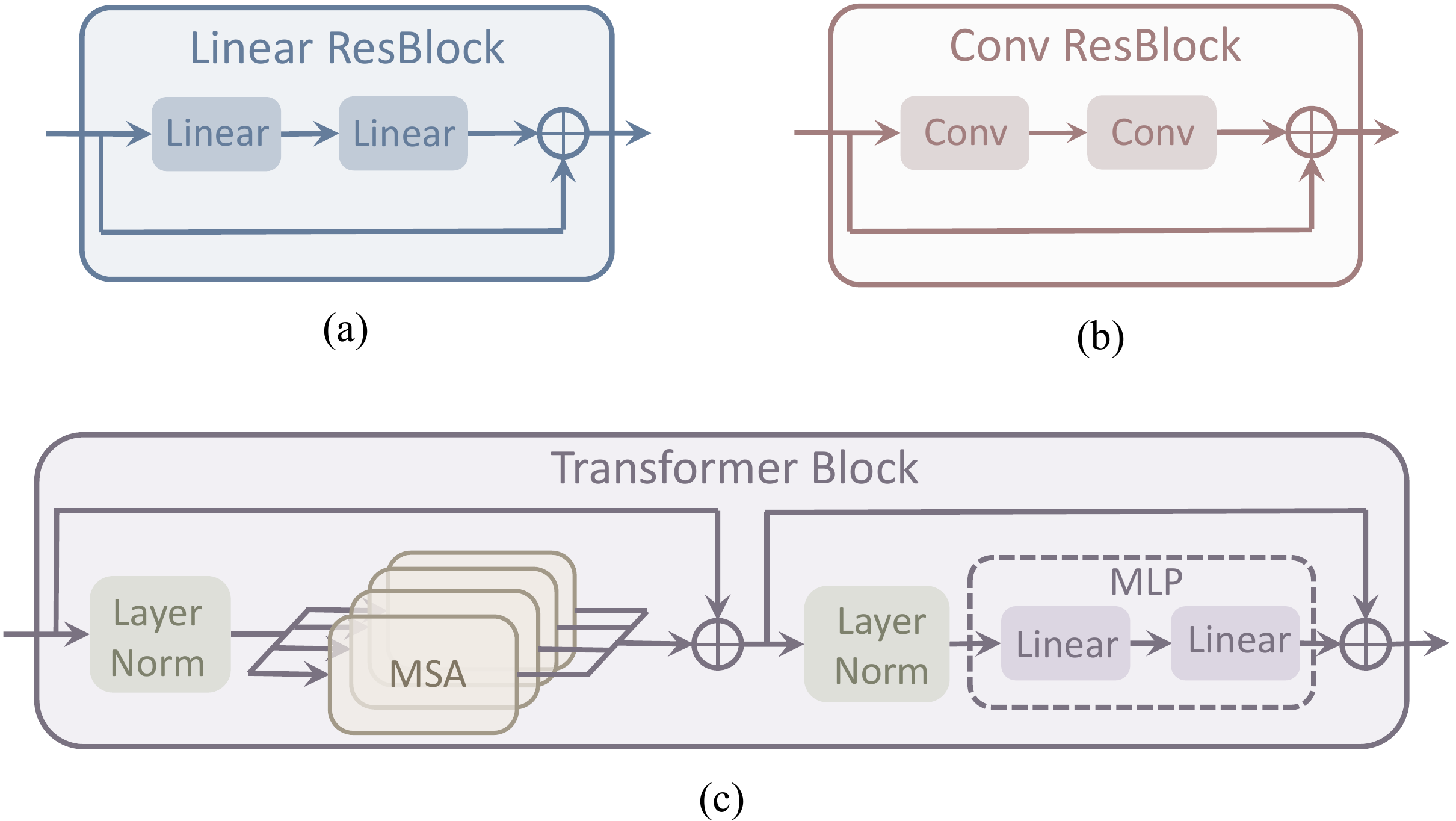} 
	\caption{Architecture of different blocks. For Linear and Conv ResBlocks, each layer is followed by a ReLU \cite{nair2010rectified} activation function. For transformer block, there is a GELU \cite{hendrycks2016gaussian} activation function between the two linear layers. MSA: Multi-head Self-Attention. MLP: Multi-Layer Perceptron.}
	\label{figure: blocks_architecture}
\end{figure}

\begin{table}[t]
	\footnotesize
	\centering
		\begin{tabular}{c|c|c|c|c|c}
			\hline
			Dataset  & $n'$ & $h$ & $D$ & $D'$ &Param. \\
			\hline 
			FFHQ \cite{karras2019style} & 30 & 8 &512 & 64 &95.0M\\
			Places2 \cite{zhou2017places} & 35 & 8 &512 & 64 &110.7M\\
            ImageNet \cite{deng2009imagenet} & 35 & 8 &1024 & 128 &441.7M\\
			\hline
		\end{tabular}
	\caption{UQ-Transformer with different model sizes for different datasets. $n'$ and $h$ are the number of transformer block and attention head. $D$ is the dimensionality of feature vectors that before and after each transformer block. $D'$ is the dimensionality of feature vector in each attention head.} 
	\label{tab: different_model_size_of_ut_transformer}
\end{table}

\section{Network Architecture}
\label{sec: network_architecture}
\subsection{Auto-Encoder}
\label{sec: auto_encoder_architecture}
For different datasets, we use P-VQVAE with the same model size, and the architecture of our default P-VQVAE is shown in \Tref{tab: architecture_of_p_vqvae}. The structure of Linear and Conv ResBlocks are shown in \Fref{figure: blocks_architecture} (a) and (b). In the paper, Section 4.3, several models are designed to show the effectiveness of different components in our method, including PUT$^{\rm conv}$, PUT$^{\rm one}$, PUT$^{\rm no\_ref}$, PUT$^{\rm qua0}$ and PUT$^{\rm tok}$. The auto-encoders in the last two models are the same with our default P-VQVAE. However, the auto-encoders in PUT$^{\rm conv}$ , PUT$^{\rm one}$ and PUT$^{\rm no\_ref}$ are different. For the auto-encoder in PUT$^{\rm conv}$ (denoted as P-VQVAE$^{\rm conv}$), all the linear layers in the encoder are replaced with convolution layers, and the input image is processed in a sliding window manner. Other modules in P-VQVAE$^{\rm conv}$ are the same with those in P-VQVAE. The architecture of encoder in P-VQVAE$^{\rm conv}$ (denoted as Conv-Enc) is shown in \Tref{tab: architecture_of_p_vqvae_conv}. The architecture of the auto-encoder in PUT$^{\rm one}$ is the same with P-VQVAE, except only one codebook $\mathbf{e}$ is used for training and testing. While for the auto-encdoer in PUT$^{\rm no\_ref}$, it can be obtained from P-VQVAE by removing the reference branch in decoder.

\subsection{Transformer}
The architecture of transformer block is depicted in \Fref{figure: blocks_architecture} (c).  There are several (denoted as $n'$) successive transformer blocks in UQ-Transformer. Within each transformer block, the input features will be enhanced by self-attention. Formally, let $\mathbf{\Bar{f}} \in \mathbb{R}^{\frac{HW}{r^2} \times D}$ be the input of transformer block.
At the $b$-th transformer block, the feature vectors are processed as:
\begin{equation}
\begin{aligned}
    \mathbf{\tilde{f}}^{b-1} & = \mathbf{\Bar{f}}^{b-1} + {\rm MSA}({\rm{LN}(\mathbf{\Bar{f}}^{b-1})}), \\
    \mathbf{\Bar{f}}^{b} & = \mathbf{\tilde{f}}^{b-1} + {\rm MLP}({\rm LN}(\mathbf{\tilde{f}}^{b-1})),
\end{aligned}
\label{eq: transformer_block}
\end{equation}
where ${\rm LN}(\cdot)$, ${\rm MLP}(\cdot)$, ${\rm MSA}(\cdot)$ denote layer normalization \cite{ba2016layer}, multi-layer perceptron and multi-head self-attention respectively. More specifically, given input $\mathbf{f}\in \mathbb{R}^{\frac{HW}{r^2} \times D}$, ${\rm MSA}(\cdot)$ could be formated as:
\begin{equation}
\begin{aligned}
    \mathbf{h}_{j} & = {\rm softmax}(\frac{(\mathbf{fw}_{\rm q}^{j})(\mathbf{fw}_{\rm k}^{j})^T}{\sqrt{D'}})(\mathbf{fw}_{\rm v}^{j}), \\
    {\rm MSA}(\mathbf{f}) & = [\mathbf{h}_0; \mathbf{h}_1;...;\mathbf{h}_{h-1}]\mathbf{w}_{\rm o},
\end{aligned}
\label{eq: multi_head_self_attention}
\end{equation}
where $h$ is the number of head, $\mathbf{w}_{\rm q}^{j}, \mathbf{w}_{\rm k}^{j}, \mathbf{w}_{\rm v}^{j} \in \mathbb{R}^{D \times D'}$, $\mathbf{w}_{\rm o} \in \mathbb{R}^{hD' \times D}$ are the learnable parameters. $[\cdot;...;\cdot]$ is the operation that concatenates the given arguments along the last dimension. By changing the values of $h, D, D'$ and $n'$, we can easily scale the size of UQ-Transformer.

We use UQ-Transformer with different model sizes for different datasets, which are shown in \Tref{tab: different_model_size_of_ut_transformer}. As a reminder, the configuration of transformers are the same with those in ICT \cite{wan2021high}.

\section{More Results}
\label{sec: more_results}
\label{sec: image_inainting_results}
\begin{figure*}[t]
	\centering
	\includegraphics[width=2\columnwidth]{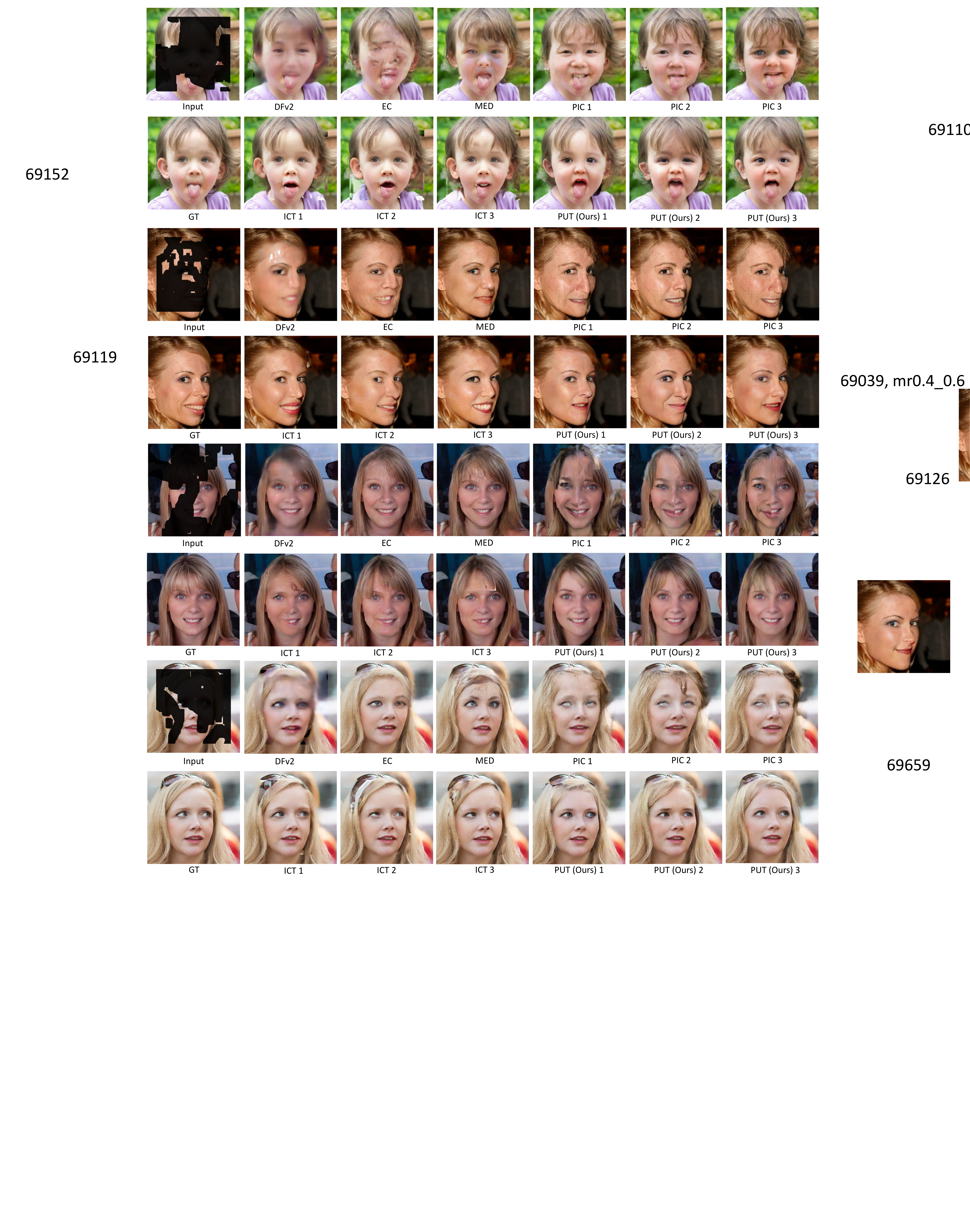} 
	\caption{Qualitative comparisons between different methods on FFHQ \cite{karras2019style}.}
	\label{figure: inpainting_results_on_ffhq}
\end{figure*}

\begin{figure*}[t]
	\centering
	\includegraphics[width=2\columnwidth]{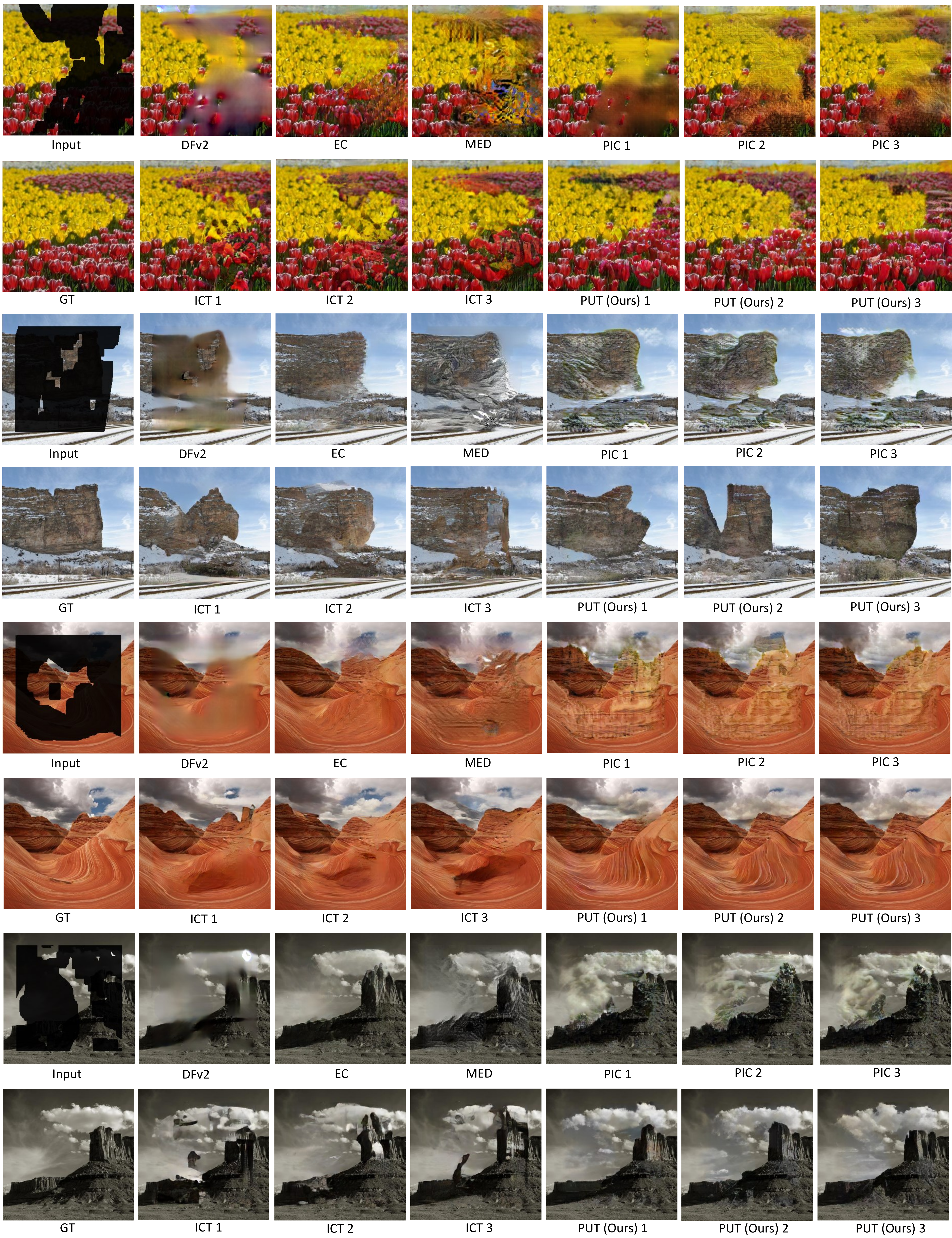} 
	\caption{Qualitative comparisons between different methods on Places2 \cite{zhou2017places}.}
	\label{figure: inpainting_results_on_naturalscene}
\end{figure*}

\begin{figure*}[t]
	\centering
	\includegraphics[width=2\columnwidth]{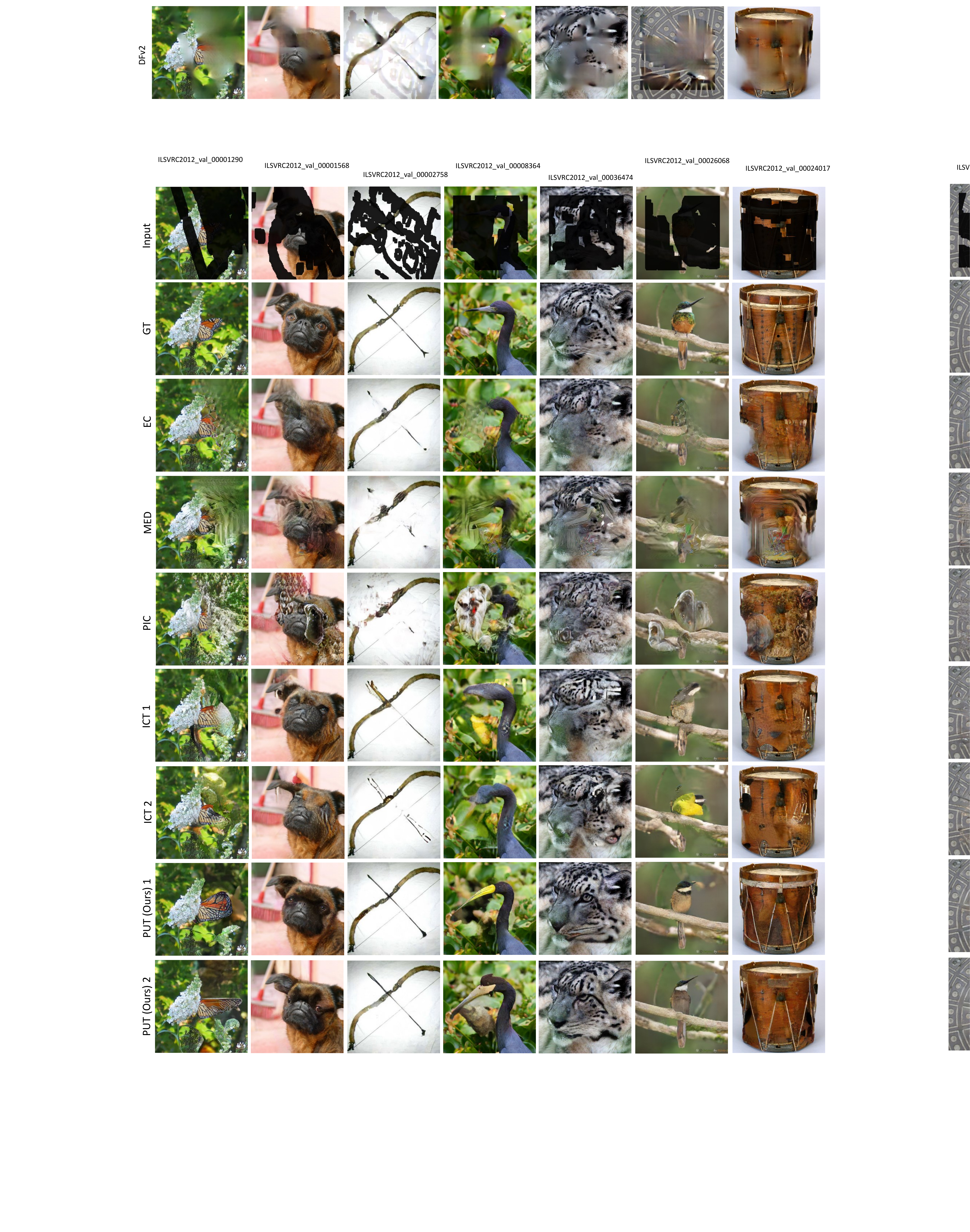} 
	\caption{Qualitative comparisons between different methods on ImageNet \cite{deng2009imagenet}.}
	\label{figure: inpainting_results_on_imagenet1}
\end{figure*}

\begin{figure*}[t]
	\centering
	\includegraphics[width=2\columnwidth]{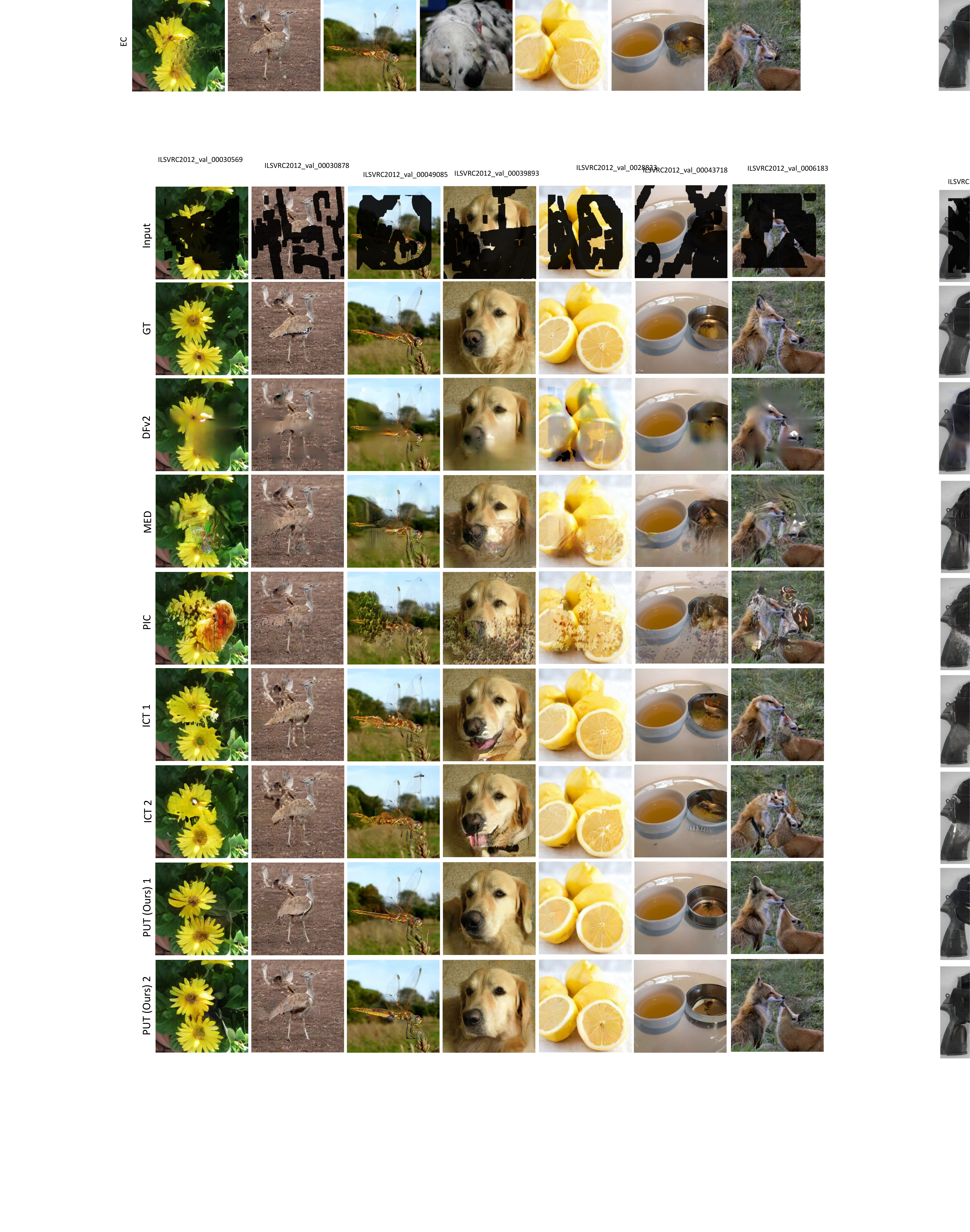} 
	\caption{Qualitative comparisons between different methods on ImageNet \cite{deng2009imagenet}.}
	\label{figure: inpainting_results_on_imagenet2}
\end{figure*}

We show more qualitative comparisons for FFHQ \cite{karras2019style} ( \Fref{figure: inpainting_results_on_ffhq}), Places2 \cite{zhou2017places} ( \Fref{figure: inpainting_results_on_naturalscene}) and ImageNet \cite{deng2009imagenet} (\Fref{figure: inpainting_results_on_imagenet1} and \Fref{figure: inpainting_results_on_imagenet2}).

\begin{table}
	\setlength{\tabcolsep}{6pt}
	\footnotesize
	\centering
		\begin{tabular}{c|c|c|c}
			\hline
			\diagbox{Models}{Datasets} & FFHQ \cite{karras2019style} & Places2  \cite{zhou2017places} & ImageNet \cite{deng2009imagenet}  \\
			\hline 
			\makecell[c]{UQ-Transformer\\(\# tokens/second)} & 37.138 & 32.048 &17.186\\
			\hline
			\makecell[c]{P-VQVAE\\(\# images/second)}  &\multicolumn{3}{c}{62.949}  \\
			\hline
		\end{tabular}
	\caption{Inference speed of different models. Tested on RTX 3090. The time consumption of P-VQVAE includes extracting feature vectors from image, quantizing feature vectors to latent vectors, and reconstructing the input image.} 
	\label{tab: inference_speed}
\end{table}

\begin{figure}[h]
	\centering
	\includegraphics[width=1.0\columnwidth]{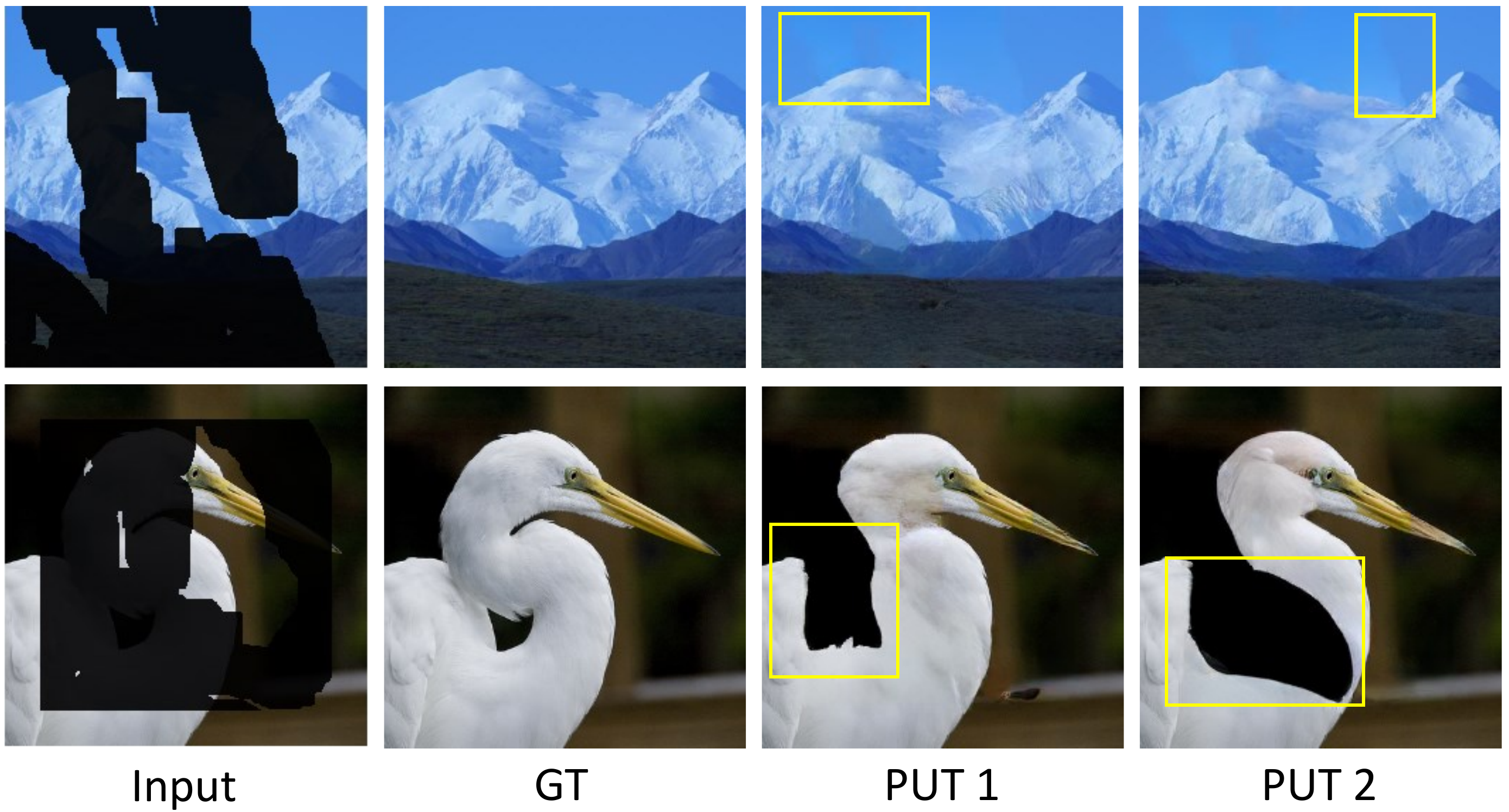} 
	\caption{Results with artifacts. Top: color distortion. Bottom: black regions. Please pay attention to the contents in yellow rectangles.}
	\label{figure: inpainting_artifacts}
\end{figure}

\section{More Discussions}
\label{sec: discussion}

\paragraph{Inference speed.} 
As mentioned in Section 5 in the paper, the main limitation of PUT is the inference speed, which is also a common issue of existing transformer-based auto-regressive methods \cite{vaswani2017attention, wan2021high, ramesh2021zero, esser2021taming}. Here we present the inference speed of PUT in \Tref{tab: inference_speed}. Note that the time consumption of inpainting a masked image depends on the area of masked regions.

\paragraph{Artifacts.} We experimentally find that there sometimes contain some artifacts in the generated results of PUT, as shown in \Fref{figure: inpainting_artifacts}. These artifacts can be divided into two categories. 1) Color distortion: the color of generated contents my not be consistent with the color of provided contents in the image. 2) Black region: PUT may produce black regions if the provided masked image contain lots of black pixels.

\end{document}